\documentclass[review,12pt,authoryear]{elsarticle}
\usepackage{amssymb}
\usepackage{amsthm}
\usepackage{amsmath}
\usepackage{times}
\usepackage{xcolor}
\usepackage[export]{adjustbox}
\usepackage{geometry}
\usepackage{caption}
\usepackage{subcaption}
\usepackage{xcolor}
\usepackage{soul}
\usepackage[usestackEOL]{stackengine}
\journal{Expert Systems with Applications}

\begin{document}

\begin{frontmatter}

\title{Self-supervised learning for hotspot detection and isolation from thermal images}

\author[inst1]{Shreyas Goyal }
\affiliation[inst1]{organization={School of Computer Science and Engineering},
            addressline={Nanyang Technological University, 50 Nanyang Avenue}, 
            postcode={639798}, 
            country={Singapore}}
\ead{shreyas007@e.ntu.edu.sg}

\author[inst1]{Jagath C. Rajapakse \corref{cor1}}
\ead{ASJagath@ntu.edu.sg}
\cortext[cor1]{Corresponding author}

\begin{abstract}
Hotspot detection using thermal imaging has recently become essential in several industrial applications, such as security applications that require identification of suspicious activities or intruders by detecting hotspots generated by human body heat, health applications such as screening of individuals in quarantine environments for onset of fever, and equipment monitoring applications where ensuring smooth operation and preventing potential malfunction by assessing temperature distribution in equipment is important. Hotspot detection is of utmost importance in industrial safety where equipment can develop anomalies. Hotspots are early indicators of such anomalies. We address the problem of hotspot detection in thermal images by proposing a self-supervised learning approach. Self-supervised learning has shown potential as a competitive alternative to their supervised learning counterparts but their application to thermography has been limited. This has been due to lack of diverse data availability, domain specific pre-trained models, standardized benchmarks, etc.
We propose a self-supervised representation learning approach followed by fine-tuning that improves detection of hotspots by classification. The SimSiam network based ensemble classifier decides whether an image contains hotspots or not. Detection of hotspots is followed by precise hotspot isolation. By doing so, we are able to provide a highly accurate and precise hotspot identification, applicable to a wide range of applications. We created a novel large thermal image dataset to address the issue of paucity of easily accessible thermal images. Our experiments with the dataset created by us and a publicly available segmentation dataset show the potential of our approach for hotspot detection and its ability to isolate hotspots with high accuracy. We achieve a Dice Coefficient of 0.736, the highest when compared with existing hotspot identification techniques. Our experiments also show self-supervised learning as a strong contender of supervised learning, providing competitive metrics for hotspot detection, with the highest accuracy of our approach being 97\%. 

\end{abstract}


\end{frontmatter}

\section{Introduction}\label{introduction}
Thermal hotspots often occur in industrial equipment, especially in electrical equipment, prior to a failure \citep{das2017finite, deng2017research}. Hotspots refer to regions that are of relatively higher temperature compared to the rest of the object or recognisable with abnormally high temperatures. It is imperative to detect hotspots in equipment and fix them early in order to prevent significant loss of financial, natural, and human resources. For instance, a hotspot or weakness in a huge tank containing oil or gas might develop into a crack and eventually lead to a complete failure \citep{galvan2013protection}. Thermal cameras (often known as infrared cameras) provide means to visualize the thermal signature of objects by processing their infrared radiations and providing us with corresponding grayscale images of this radiation.
Therefore, our objective is to undertake the task of automatic classification of images of such equipment as 'anomalous' (presence of hotspots) or 'normal'. Attempts would also be made to pin-point the location of the hotspots once the image is classified as anomalous in order to make swift remedial action possible. In this work, we propose a self-supervised deep learning approach for hotspot detection from thermal images. 

Traditional methods of hotspot detection are done by supervised techniques that are dependent on huge amounts of labelled images \citep{winston2021solar, ali2020machine, chen2020hot, dit2018localization, herrmann2016real}. However, such large-sized labelled thermal image datasets are hard to come by for different application requirements as image dataset labelling is a time-consuming and expensive process. Furthermore, in the case of industrial equipment, the situation is even worse given the lack of appropriate thermal image datasets for the same. Hence, we resort to self-supervised learning, a training algorithm that learns data characteristics through unlabelled data. This makes the method malleable for a variety of similar applications. Self-supervised learning also mitigates the dependence on time and effort involved in labelling image datasets \citep{newell2020useful}. Moreover, the proposed approach is suitable to be used as a general initial confirmation on the presence of an anomaly in a particular region as well. This could then be followed by downstream tasks to pin-point the anomalies for further processing or rectification.

Given the paucity of thermal image datasets in general, we have created a new thermal image dataset with different shaped metallic sheets as image objects. This dataset will be made freely available on request at: https://github.com/SCSE-Biomedical-Computing-Group/Thermal-Images-for-Hotspot-Detection-TIHD-Dataset.

\subsection{Contributions of this work:}
\begin{enumerate}
  \item We are the first to employ a self-supervised algorithm for hotspot detection in thermal images.
  \item To enhance the performance of our approach, we made modifications to the SimSiam backbone: (i) by replacing it with the XceptionNet model and (ii) by adding a cross-entropy loss term to the loss function. We introduced an intermediate step of image classification in the hotspot detection pipeline by ensembling two encoders from different SimSiam models in order to minimize false positives in day-to-day use of this technique, which helps by filtering out false positives and thereby improving the robustness of the hotspot detection process.
  \item We combined the contrastive learning method, the classification results of the downstream task, and class-activation maps to come up with a novel thermal image deep-learning hotspot isolation method.
  \item We created a novel high-quality thermal image dataset for hotspot detection to address the issue of lack of appropriate thermal image datasets for the same.
\end{enumerate}

\subsection{Related Work}\label{related-work}
Both supervised and unsupervised techniques have been proposed for thermal image hotspot detection. K-Means \citep{macqueen1965some} clustering has been the most popular technique for hotspot detection.
In \cite{mohd2017application} K-Means clustering algorithm was used for segmenting a thermal image into the foreground and background, with the expectation that the hotspot would get separated from the foreground. This way, they were able to limit the amount of image pixels for subsequent processing. However, before the segmentation, the images were converted into the CIE L*a*b* colour space since L*a*b includes all perceivable colours. Mohd et al. conducted tests on a number of images but they have not mentioned the nature of the targets in the images. However, they were successfully able to remove the unassociated background through segmentation, and thereby focused on a smaller region containing the hotspot. However, using this method the area containing the hotspot obtained through segmentation overestimated the area of the hotspot. This was especially the case with the existing technique when multiple hotspots were present in the subject.   
In \cite{salazar2016hotspots} a K-Means thermal image segmentation technique was also used for finding and isolating hotspots in solar cells, for which they developed a 'Hotspot Detection Program' . 
From the examples provided we assume that the algorithm works well on hotspots having a large size. Furthermore, the technique requires a lot of manual input by the user.
A fixed $k$ value of 3 was used for the purpose of segmentation. The user needs to specify the temperature range of the uploaded image for the hotspot average temperature function to work. The user is also supposed to manually select a bounding box around the relevant parts of the image.  

Afifah et al. used a multilevel Otsu segmentation algorithm for isolation of hotspots from thermal images of photovoltaic cells \citep{afifah2021new}. This work came as an improvement to their earlier work based on simple Otsu segmentation \citep{afifah2020hotspot}.  
The methodology consisted of taking multiple thresholds for segmenting the image and then utilizing the most significant threshold (largest) to binarize the original image. Afifah et al. utilized morphological opening to get rid of noise in the final binary image. 
A comparison between the K-means segmentation algorithm and the multilevel Otsu segmentation algorithm for the detection of hotspots in thermal images was done in \cite{lim2022automatic}. For both these algorithms, they divided the image into six segments and then calculated the average temperature of each segment to find the one with the hotspot based on which segment had the highest temperature. They found that the Otsu segmentation algorithm provided higher accuracy and had performed faster than the K-means segmentation algorithm. This technique can be considered to be more robust than the one proposed by \cite{afifah2021new} since it relies on calculating the average temperature of each segment after segmentation for detecting the hotspots, rather than converting the image into a binary image based on the largest hotspot.

\cite{alajmi2019ir} proposed a hotspot detection technique for photovoltaic cells using thermal images. They converted the thermal image to grayscale and applied edge preservation and smoothening filters. The processed image was then converted to the HSV space, using which a binary mask for each channel got created. The use of specific lower and upper thresholds to obtain the masks for each of the three channels was proposed.  
It is possible that the specific thresholds proposed in this paper do not work well to isolate the hotspots in all thermal images since the colours displayed in a thermal image is relative to the surrounding temperatures. This means that a particular colour can represent different temperatures in different images and therefore the thresholds might be different for different images. 

\cite{ali2020machine} proposed a hotspot detection and classification approach for classifying thermal images of PV panels as 'healthy', 'hotspots', or 'faulty' using a hybrid features approach for dataset construction, along with an SVM classifier. The proposed features consisted of HOG, LBP, RGB, contrast, correlation and energy descriptors. Following this, \cite{ali2022early} proposed another PV thermal image classification technique using machine learning methods on colour image descriptors as features. Even though they did not attempt hotspot isolation, they classified images as normal, having hotspots, or as completely defective with high accuracy. Ali et al. carried out experiments with many image colour descriptors like histograms, SIFT methods, and colour moments, and got the highest classification accuracy using the k-Nearest Neighbours method with the rgSIFT descriptor on an image size of 71x71 pixels. Ahmed et al. extracted thermal image features using the SURF method, and utilized the K-Means clustering algorithm for creating a visual vocabulary. Shallow classifiers like SVM, Naive Bayes, k-Nearest Neighbours are then utilized with this vocabulary for classifying images of photovoltaic panels as healthy, hotspot, or faulty  \citep{ahmed2022visual}. Following this work, Ahmed et al. merged the benefits of pre-trained deep learning models like ResNet18, SqueezeNet, and GoogleNet through transfer learning and of shallow classifiers like SVM, k-Nearest Neighbours, naive Bayes to classify thermal images of photovoltaic panels as healthy or as one of the five different types of defects \citep{ahmed2023comparison}. In \cite{accikgoz2022classification} different deep learning models like ResNet50, AlexNet, SqueezeNet, GoogLeNet, ShuffleNet and MobileNet were compared for classifying thermal images of solar panels as 'Hotspot' or 'No anomaly' depending on whether the image had a hotspot or not. They found that AlexNet and MobileNet had the best performance on different metrics. In \cite{songbosf} the Bag-of-Features model(BOF) was integrated with the SURF technique to come up with a Bag-of-SURF-Features (BOSF) vector for each image. They used these features in a SVM for classification of two types of circuit board images, into faults that lead to hotspots in circuit boards. Song et al. showed that the BOSF feature extraction technique performed better than other feature extraction techniques, and that the SVM classifier performed better than a Naive Bayes, or an error back propagation neural network.

From these works it was observed that a lot of the existing hotspot detection techniques are not precise in pinpointing hotspots, require manual inputs at times, and lack robustness and generalizability in their application. Furthermore, most of the existing techniques do not deal with hotspot isolation at all. Therefore, in order to tackle these issues and make the hotspot detection process reliable, we propose a self-supervised learning hotspot detection approach.

A few recent works that utilize self-supervised and semi-supervised learning techniques for classification have been proposed for different applications.
In \cite{montanaro2022semi}, the task of land cover classification through the use of multi-modal image data with the help of self-supervised contextual learning methods was undertaken.  
The approach involved a preprocessing step that dealt with feature extraction, followed by two steps of self-supervised learning. The first self-supervised step’s goal was to reduce the distance between features from the same classes of different modalities and the goal of the second self-supervised step was to enable learning of pertinent features for facilitating land-cover classification.
In \cite{xu2022exploring}, a modified Simple Siamese model for classifying ships using SAR (Synthetic Aperture Radar) images was proposed. They took into account polarized images in addition to regular augmentations as positive samples. Xu et al. also compared different similarity functions and achieved an accuracy of 66\% by using this approach. However, self-supervised methods have not been employed for hotspots detection. 

This article has been organised as follows: This introduction is followed by a section on the methods (Section \ref{methods}) used for the development of the proposed approach. This is followed by a section on the experiments and results (Section \ref{results}) obtained on different datasets and metrics using the developed approach. Lastly, a brief section on discussion and conclusions (Section \ref{conclusions}) that can be derived from the performed experiments and corresponding results is presented as well.

\section{Methods}\label{methods}
 In order to make the anomaly detection process robust and due to a lack of suitable annotated data for training, a self-supervised classification strategy is developed in this work. Self-supervised learning and subsequent fine-tuning for downstream tasks are limited in applications to thermal image domain.

\subsection{Self-supervised learning}
Self-supervision allows networks to learn from unlabelled data. Its beauty lies in its ability to leverage on the immense amount of unlabelled data that is always available but are expensive and time consuming to label. Self-supervised learning overcomes this limitation by solving a proxy or surrogate task that requires just the availability of unlabelled data. With these proxy tasks, the model learns to predict certain aspects of the data without any supervision or annotations.
The proxy tasks are designed to provide the model with a different type of supervised learning paradigm in which just the unlabelled data are utilized. Usually, some kind of data-augmentation or transformation of the unlabelled data is undertaken so as to generate pseudo-labels. This can be thought of as withholding or distorting a part of the data. Given this augmented or transformed data, the network is then tasked with reconstructing the original image with the original image or some part of it, thus behaving like pseudo-labels. Some examples of proxy tasks include image in-painting, image colourization, rotation prediction, etc. Once the network is successfully able to reconstruct the original image to a close degree of similarity, it can be understood that the model has been able to learn a rich and meaningful representation of the data.

A common approach in self-supervised learning attempts to learn a meaningful representation from unlabelled data by contrasting between differently augmented views of the same image. This means that two different augmentations from the same set of data augmentations are applied on an image before feeding the pair to the model. This is known as \textit{contrastive learning}. The aim in this case is to learn an embedding space in which alike samples (different augmentations of the same image) get pushed close to each other whereas different samples get pushed apart. This allows the model to learn discriminative representations. The knowledge of the dataset that the model gains in this process can then be utilized in an efficient manner for different downstream tasks like classification, detection, etc. Some of the common losses used for contrastive learning include contrastive loss \citep{chopra2005learning}, triplet loss \citep{schroff2015facenet}, and the InfoNCE loss \citep{oord2018representation}.

In summary, the beauty of self-supervised learning lies in its ability to unlock and leverage the potential of unlabelled data, allowing the models to learn meaningful representations of the data. The methodology requires designing a proxy task to guide the learning process, leveraging the structures within the data. Once the model is able to learn a rich representation of the unlabelled data, it can be transferred to a downstream task like classification, segmentation, detection etc. where it can be fine-tuned using a smaller labelled dataset for the specific application. Thereby reducing dependency on labelled data.

One of the popular contrastive learning framework is SimSiam (Simple Siamese) representation learning \citep{chen2021exploring}. It consists of a shared encoder that receives two augmentations or views ($v_1$ and $v_2$) of an image, and a predictor that transforms one of the encoded views to match the encoding of the other view. The network learns through a loss by matching a transformation of this encoded view of the image to the encoding of a different view of the same image. This means that an attempt is made by the model to minimize the distance between the different augmentations of the same image. In this process, the encoder of the first view is able to learn a meaningful data representation that can be fine-tuned and transferred to several downstream tasks like classification, detection, etc.
 
\begin{figure}[ht]
    \centering
    \begin{subfigure}{0.32\textwidth}
        \centering
        \includegraphics[width=0.75\textwidth]{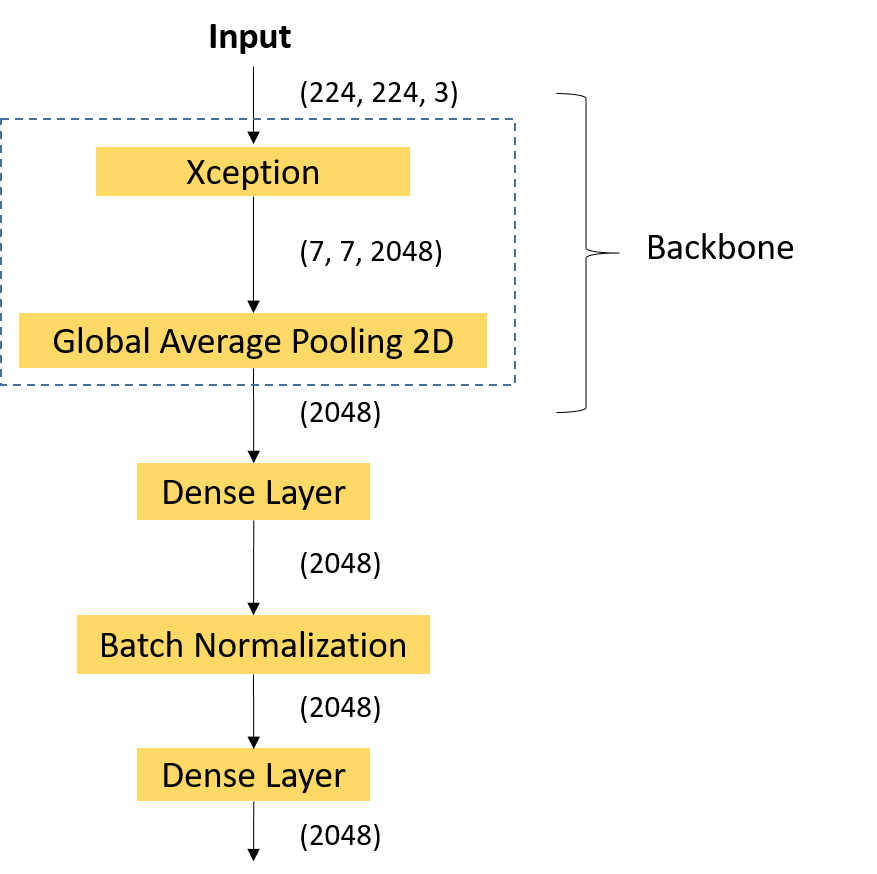}
        \caption{}
        \label{fig:encoder}
    \end{subfigure}
    \begin{subfigure}{0.32\textwidth}
        \centering
        \includegraphics[width=0.65\textwidth]{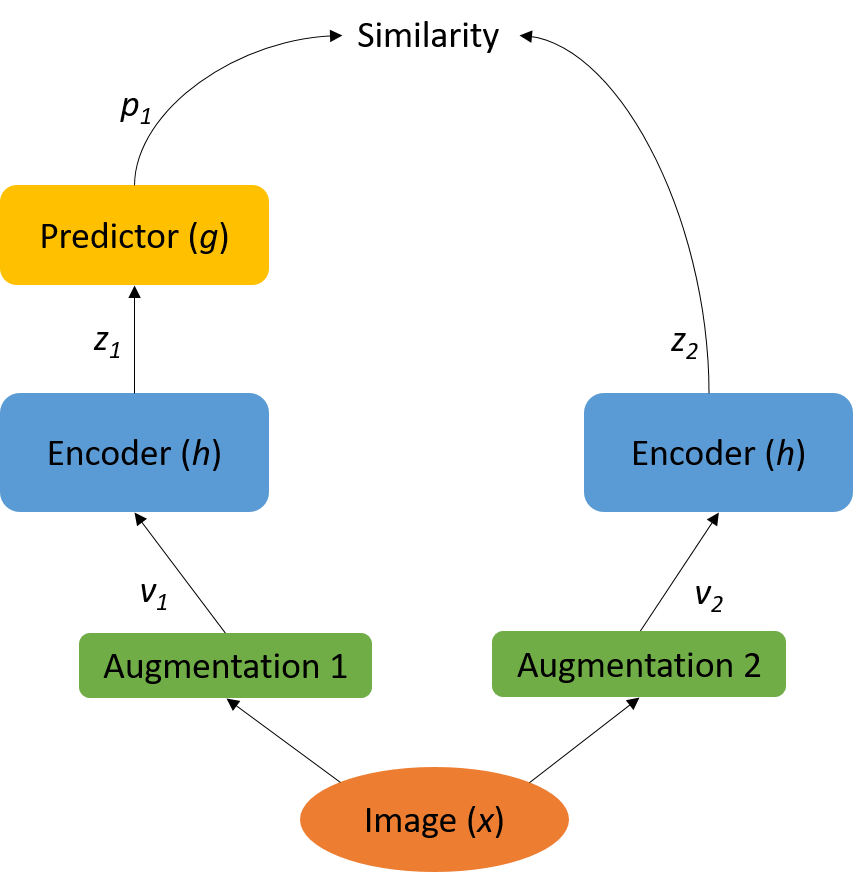}
        \caption{}
        \label{fig:ourssl}
    \end{subfigure}
    \begin{subfigure}{0.32\textwidth}
        \centering
        \includegraphics[width=0.45\textwidth]{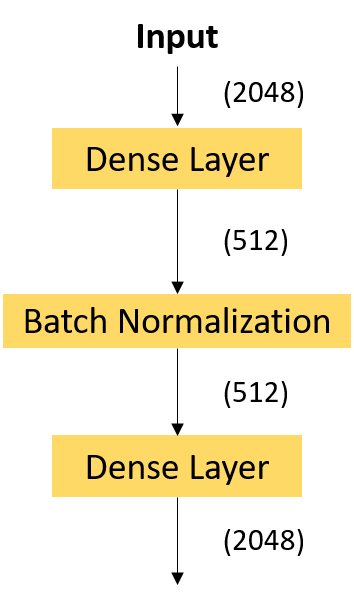}
        \caption{}
        \label{fig:predictor}
    \end{subfigure}\\
    \caption{(a) Contrastive learning encoder architecture with Xception backbone (b) Illustration of self-supervised learning of the encoder; the encoder is shared for processing of the two views $v_1$ and $v_2$ of input image $x$, and (c) The predictor architecture (MLP head) in the contrastive learning framework.}
    \label{fig:fulloverview}
\end{figure}

 Given the paucity of labelled images, the importance of hotspot detection and isolation, and due to the competitive performance of self-supervised methods \citep{jaiswal2020survey, chaves2023evaluation}, we propose a self-supervised contrastive learning approach for the identification of thermal images containing hotspot anomalies through classification. For this purpose, we propose a process that utilizes a modified SimSiam framework, in which an XceptionNet \citep{chollet2017xception} is used instead of the ResNet50, the loss function is modified to include a cross-entropy function, and an ensemble of two encoders from two independently trained SimSiam frameworks is used for classification. The illustration of our model overview can be seen in Figure \ref{fig:fulloverview}. SimSiam was picked as the framework of choice among other frameworks like SimCLR \citep{chen2020simple}, MoCo \citep{he2020momentum} and BYOL \citep{grill2020bootstrap} because in the case of SimSiam, there is no need for a momentum encoder, large batch sizes, or negative samples because of its use of a stop gradient. These benefits further improve the training duration and require significantly fewer computing resources. Therefore, a SimSiam framework was used to train on a large number of unlabelled image dataset to let the model learn a good representation of the data. Following this, the encoder of the model was fine-tuned using a labelled dataset as appropriate for the application.

\subsection{Model architecture}
\subsubsection{Encoder and Predictor}
Our self-supervised learning framework consists of a shared encoder (consisting of the backbone network) and a predictor as shown in Figure \ref{fig:ourssl}. An Xception network architecture without its head is used as the backbone to learn the representations of the unlabelled infrared images. For coming up with a suitable backbone network, we conducted a thorough comparison of various convolution neural networks to determine the most suitable candidate. The compared models included ResNet50 \citep{he2016deep}, VGG16 \citep{simonyan2014very}, XceptionNet, and EfficientNet \citep{tan2019efficientnet}. Among these models, the XceptionNet demonstrated notably most stable training behaviour compared to the counterparts. The training process of XceptionNet exhibited minimal fluctuations in loss and accuracy on its way to convergence. Given this, the XceptionNet provided a reliable foundation for further experimentation and was therefore chosen as the backbone of choice over ResNet50 used in \cite{chen2021exploring}. The same point will be further established in the fourth step of the ablation study in Section \ref{ablation}.

The output of the encoder is obtained by applying a 2D global average pooling on the output of the Xception network (backbone). This is followed by applying a dense layer, batch normalization, and a dense layer. The output of the encoder so formed is then sent to the predictor which consists of an multilayer perceptron (MLP) head. The architecture used for the Xception network was the same as its original description \citep{chollet2017xception}. The final encoder network architecture is shown in Figure \ref{fig:encoder}.

The predictor MLP head takes the output of the encoder as its input. This means that it takes a 2048 dimensional input on which a dense layer is applied. This is followed by a batch normalization and a dense layer. The predictor MLP head architecture is shown in Figure \ref{fig:predictor}.

The encoder learns representations of the thermal images fed to it by processing two different augmentations, $v_1$ and $v_2$, of the same image input $x$ (Figure \ref{fig:ourssl}). These differently augmented images are considered as positive (similar) samples. It processes these two augmentations simultaneously through weight sharing. The projection outputted by the encoder for one of the views is fed to the predictor, architecture of which is given in Figure \ref{fig:predictor}. The predicted output for the predictor-encoder branch for the view $v_1$ is $p_1 = g(h(v_1))$ where $h$ is the encoding function and $g$ is the predicting function. The framework then matches this prediction to the encoder projection of the other view $v_2$ of the same image: $z_2 = h(v_2)$ (i.e., encoder only branch) by minimising the negative of the dot product between $L_2$ normalized prediction and projection (negative cosine similarity) of the same image. Since the two samples are considered as positive, the predictor's output serves as a target for the encoder. On the encoder only branch, a stop-gradient is used while training, meaning that the projection of the view is treated as a constant and the corresponding encoder does not receive any gradient backward for updating. 

The gradient backpropagation is stopped in the encoder only branch to prevent learning degeneration and leads to reasonable representation learning even without negative samples.
The negative cosine similarity between the projections of the two views is given by:
\begin{equation}\label{eq:cosinesimilarity}
    D(p_{1}, z_{2}) = - \frac{p_{1}}{||p_{1}||_{2}}.\frac{z_{2}}{||z_{2}||_{2}}
\end{equation}
where $p_{1}$ is to the prediction of the projection of view $v_1$ of the image and $z_{2}$ is to the projection of the other view $v_2$ of the same image. $\| . \|$ denotes $l_2$ norm.

The gradient is stopped from propagation on the encoder only branch during training. This leads to the encoder of that particular view not receiving any gradient from the projection. The objective function is derived by using the similarity in (\ref{eq:cosinesimilarity}) and includes the stopped gradient $D(p_{1}, sg(z_{2}))$ where $sg()$ denotes that the argument inside is considered as a constant when computing the gradient. On the other hand, the encoder and the predictor get trained during the training process in the branch on which the gradient propagates. A symmetric version of the cosine similarity objective function shown above is used in practice. The final cosine similarity objective function is written as:
\begin{equation}\label{eq:finalcosinesimilarity}
    L_{1} = D(p_{1}, sg(z_{2})) + D(p_{2}, sg(z_{1}))
\end{equation}

In our model, we modified above loss function by adding the cross-entropy loss function (\ref{eq:cross-entropy}) in order to bring in some amount of regularization in order to prevent overfitting. It was also expected that introducing the cross-entropy loss would help the model traverse multiple alternative solutions in the latent space. Finally, the additional cross-entropy loss term provides further discrimination power to the model in order to further improve down stream classification accuracy. The cross-entropy loss is given by
\begin{equation}\label{eq:cross-entropy}
    H(p1, z2) = - \sum p_{1}log(z_{2})
\end{equation}
The final loss equation is given by
\begin{equation}\label{eq:finalloss}
     L = D(p_{1}, sg(z_{2})) + D(p_{2}, sg(z_{1})) + \beta \Bigl(H(p_{1}, sg(z_{2})) + H(p_{2}, sg(z_{1}))\Bigr)
\end{equation}
Again, a symmetric version of the cross-entropy loss was used to train the model, and it was added to the cosine similarity loss equation (\ref{eq:finalcosinesimilarity}) along with weight ($\beta$) for the cross-entropy loss. The weight was determined empirically in our experiments (Section \ref{trainingencandpred}). 

\subsubsection{Data Augmentation}
\begin{figure}
    \centering
      \includegraphics[width=0.4\textwidth]{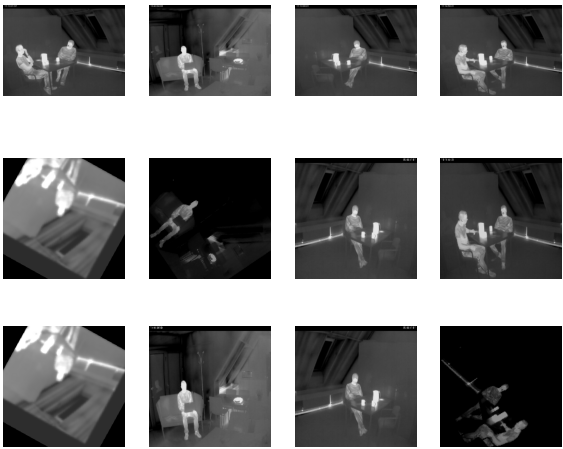}
    \caption{Augmentation examples. The first row shows the original images whereas the second and the third rows show the two augmentations that have been applied to the original image in the corresponding column.}
    \label{fig:augmentation_samples}
\end{figure}

Each image used for training was augmented twice to create two different views of each image for inputting into the network. The augmentations were achieved by image translation, left-right flip, Gaussian blurring, colour jittering, colour dropping and rotation. A random amount of each of these augmentations were applied to each image twice. Prior to data augmentation, the image size was decreased to 224x224x3. A few examples of the augmentation are shown in Figure \ref{fig:augmentation_samples}.

\subsection{Hotspot detection}
\begin{figure}
    \centering
    \includegraphics[width=0.27\textwidth]{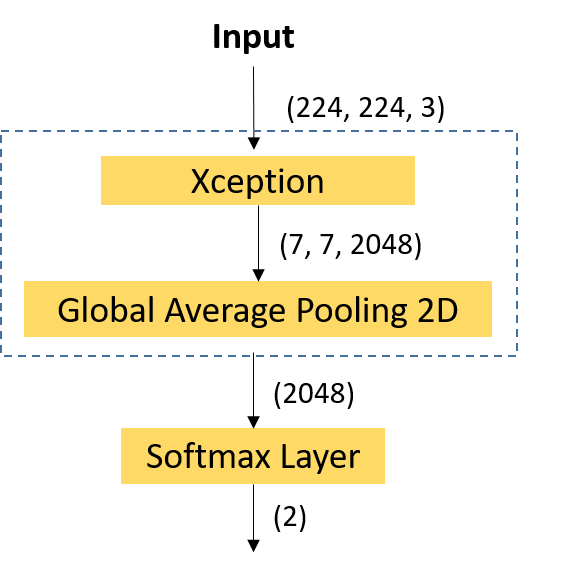}
    \caption{Hotspot detection architecture.}
    \label{fig:classifier}
\end{figure}

Once the training with the objective function in (\ref{eq:finalloss}) is complete, the trained encoder is transferred to the downstream classification task. For this, a classification network was created using the weights of the trained encoders followed by a softmax layer, as shown in Figure \ref{fig:classifier}. A labelled dataset consisting of labelled 'anomalous' and 'normal' images was used for tuning the classification network. The classifier consists of the trained encoder with an intermediate layer for the purpose of extraction of image features. This is followed by a dense layer with two units having softmax activation function since this is a 2-way classification task. The loss function used for fine-tuning is the sparse categorical cross-entropy loss and the Adam optimizer was used for learning. Following the completion of the fine-tuning process, this classifier is used for classifying images as 'normal' or 'anomalous'. For our proposal, two SimSiam frameworks were trained, and an ensemble of two encoders from these models was used for classification. One of the SimSiam framework training consisted of training with the original loss function in (\ref{eq:finalcosinesimilarity}), and the training dataset, whereas the other framework was trained with the compound loss in (\ref{eq:finalloss}), and the training dataset. This was done in order to utilize the benefits of both the independently and uniquely trained SimSiam frameworks while carrying out the downstream classification task. The ensemble network was limited to two models to control the training time, and the number of parameters. Further details are presented in Section \ref{results}.

\subsection{Hotspot isolation}
In order to precisely identify the hotspots and further verify the working of our classifier, and for the purpose of model explainability, GradCAM \citep{selvaraju2017grad} was utilized to visualize the importance maps generated by our network and then isolate the hotspots.

\begin{figure}
    \centering
    \begin{subfigure}{0.30\textwidth}
        \centering
        \includegraphics[width=0.79\textwidth]{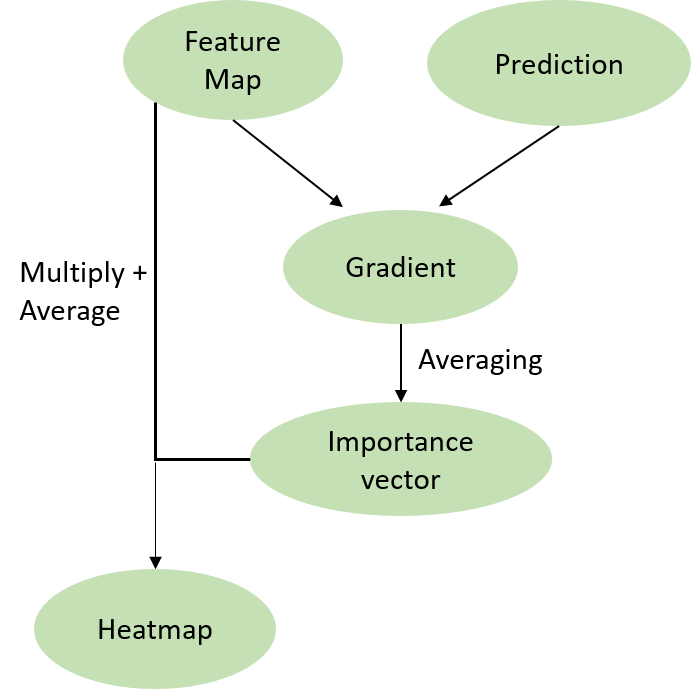}
        \caption{}
        \label{fig:gradcam}
    \end{subfigure}
    \begin{subfigure}{0.30\textwidth}
        \centering
        \includegraphics[scale = 0.07]{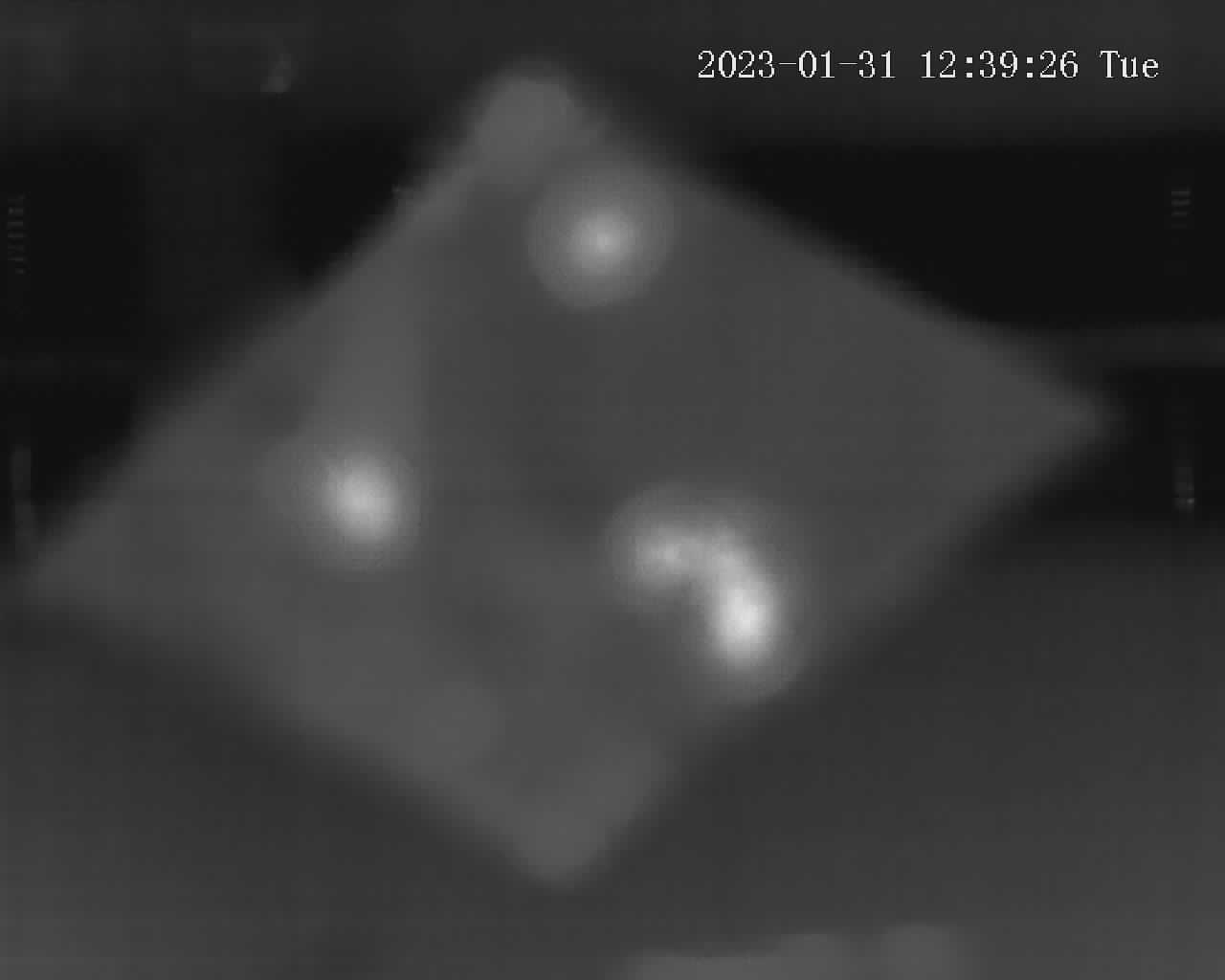}
        \includegraphics[scale = 0.07]{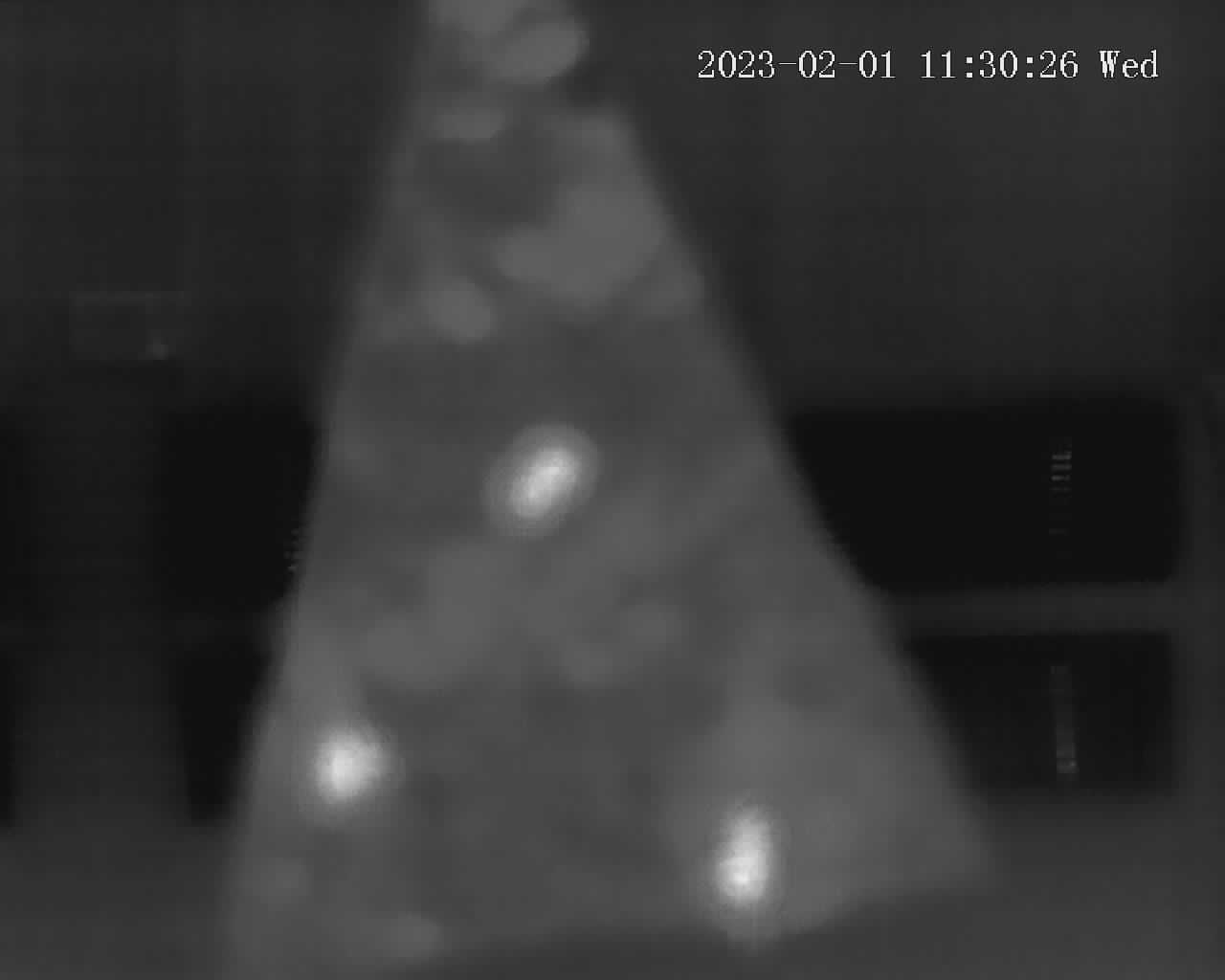}
        \caption{}
        \label{fig:Grad-CAM 2.4}
    \end{subfigure}
    \begin{subfigure}{0.30\textwidth}
        \centering
        \includegraphics[scale = 0.07]{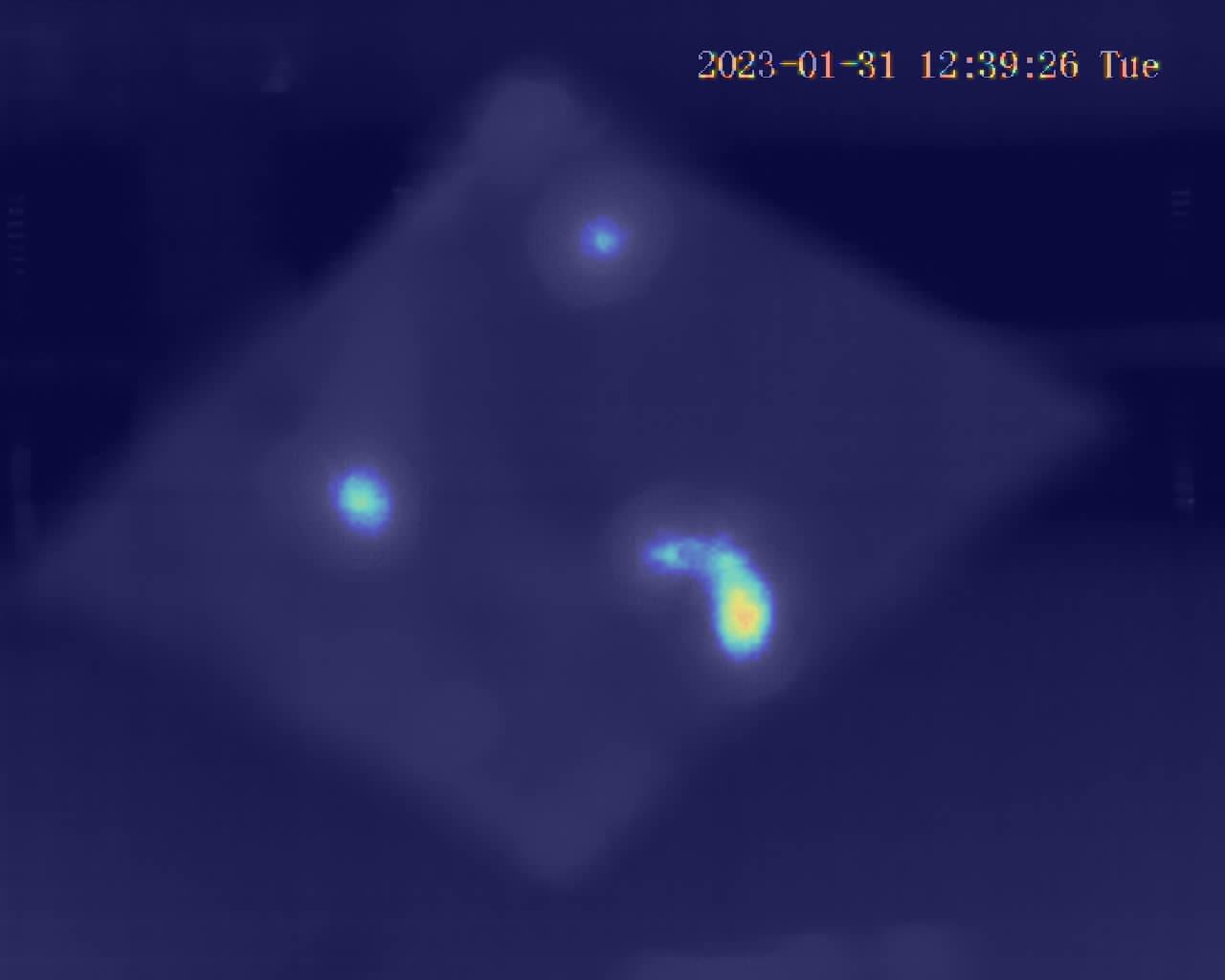}
        \includegraphics[scale = 0.07]{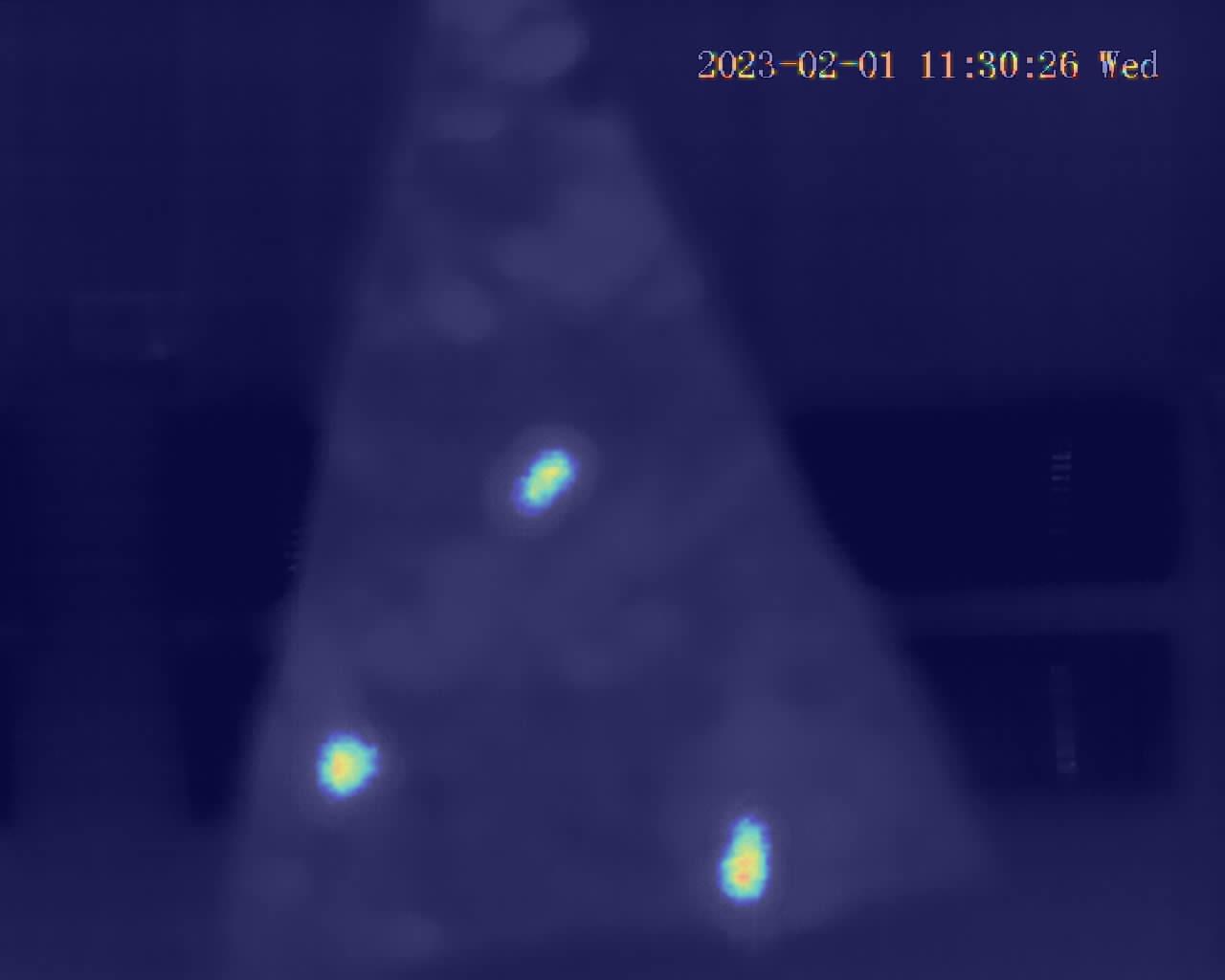}
        \caption{}
        \label{fig:Grad-CAM 2.4_}
    \end{subfigure}
    
    \caption{(a) Hotspot isolation process overview, (b) Sample input images from the TIHD dataset, (c) GradCAM output images.}
    \label{fig:isolation_overview and samples}
\end{figure}

This was done by creating a new model, taking the same inputs as our Xception network but truncating it to the final convolution layer (final layer outputting a 4 dimensional output). Therefore, this model outputs a batch of features of the input image. This is followed by calculating the gradient of the prediction made by the classifier model with respect to the feature map outputted by the final convolution layer of the classifier. The gradients of the different feature maps were averaged (summarized). This outputs a vector with the values representing the mean intensity of each feature map channel. These values represent the importance of each channel with respect to the particular prediction made by the classifier. Therefore, in order to find the importance heatmap, these mean values of the gradient of each feature map is multiplied with the original feature map channels. This would tune the values of each feature map channel to its relative importance. 
Finally, in order to obtain the heatmap, the feature volume obtained from the previous step of multiplying the gradients and the original feature map is averaged along the channels dimension to get a 2D resulting heatmap. The heatmap is then normalized for ease of visualization. The heatmap allows us to see where in the input image, the classifier focuses while making its predictions. Therefore, if the classifier has been trained in a reasonable manner, the heatmaps generated in this manner would capture the hotspots specifically, while ignoring the other elements in the image. This way, the application of GradCAM in our model was furthered by utilizing it for pinpoint or isolate hotspots directly from the importance heatmap outputted by it. The overview of this method is given in Figure \ref{fig:gradcam}. A few sample images showing the output of the GradCAM approach are also given in Figures \ref{fig:Grad-CAM 2.4} and \ref{fig:Grad-CAM 2.4_}.

\section{Experiments and Results}\label{results}
All the experiments were conducted on a server with one Nvidia A100 GPU. The codes were written in Python 3 and using Tensorflow 2.0 library.

\subsection{Datasets}
Self-supervised learning models were trained and fine-tuned on two datasets: AAU VAP Trimodal dataset and the Thermal Images for Hotspot Detection (TIHD) dataset.

\subsubsection{AAU VAP Trimodal Dataset}
\begin{figure}
    \centering
    \begin{subfigure}{\textwidth}
        \centering
        \includegraphics[scale=0.17]{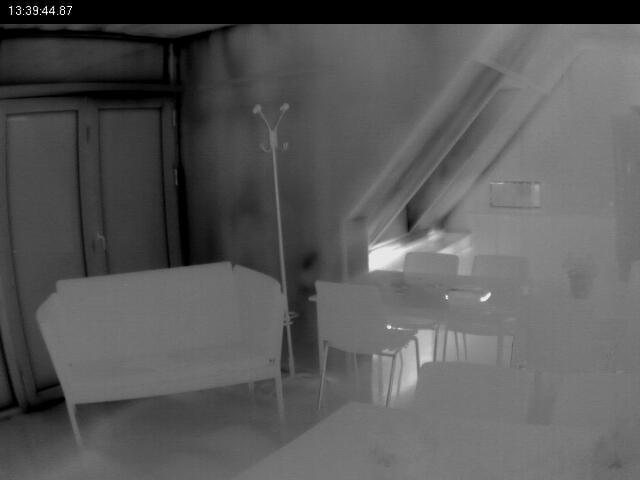}
        \includegraphics[scale=0.17]{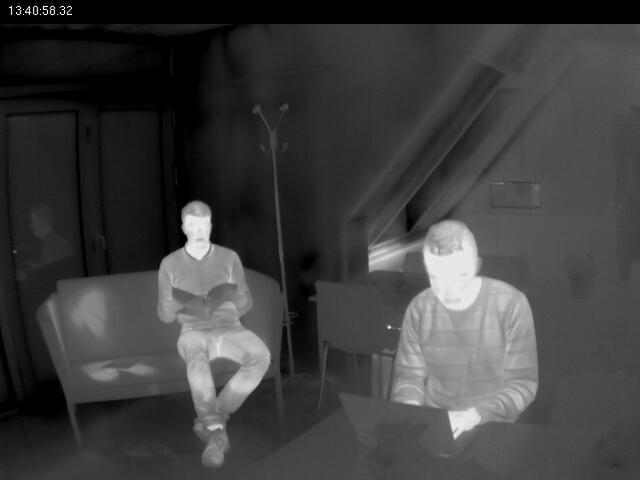}
        \includegraphics[scale=0.17]{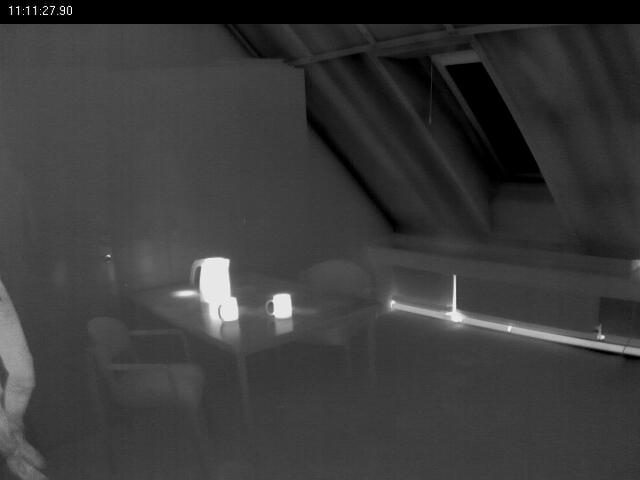}
        \includegraphics[scale=0.17]{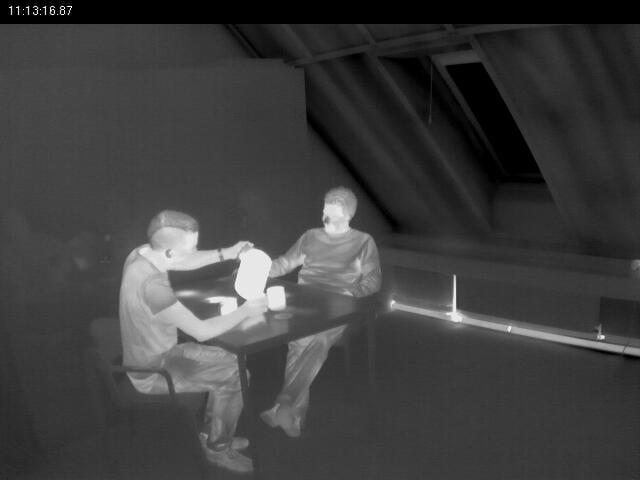}
        \caption{}
        \label{fig:proxy_samples}
    \end{subfigure}\par\medskip
    \begin{subfigure}{\textwidth}
        \centering
        \includegraphics[scale=0.0847]{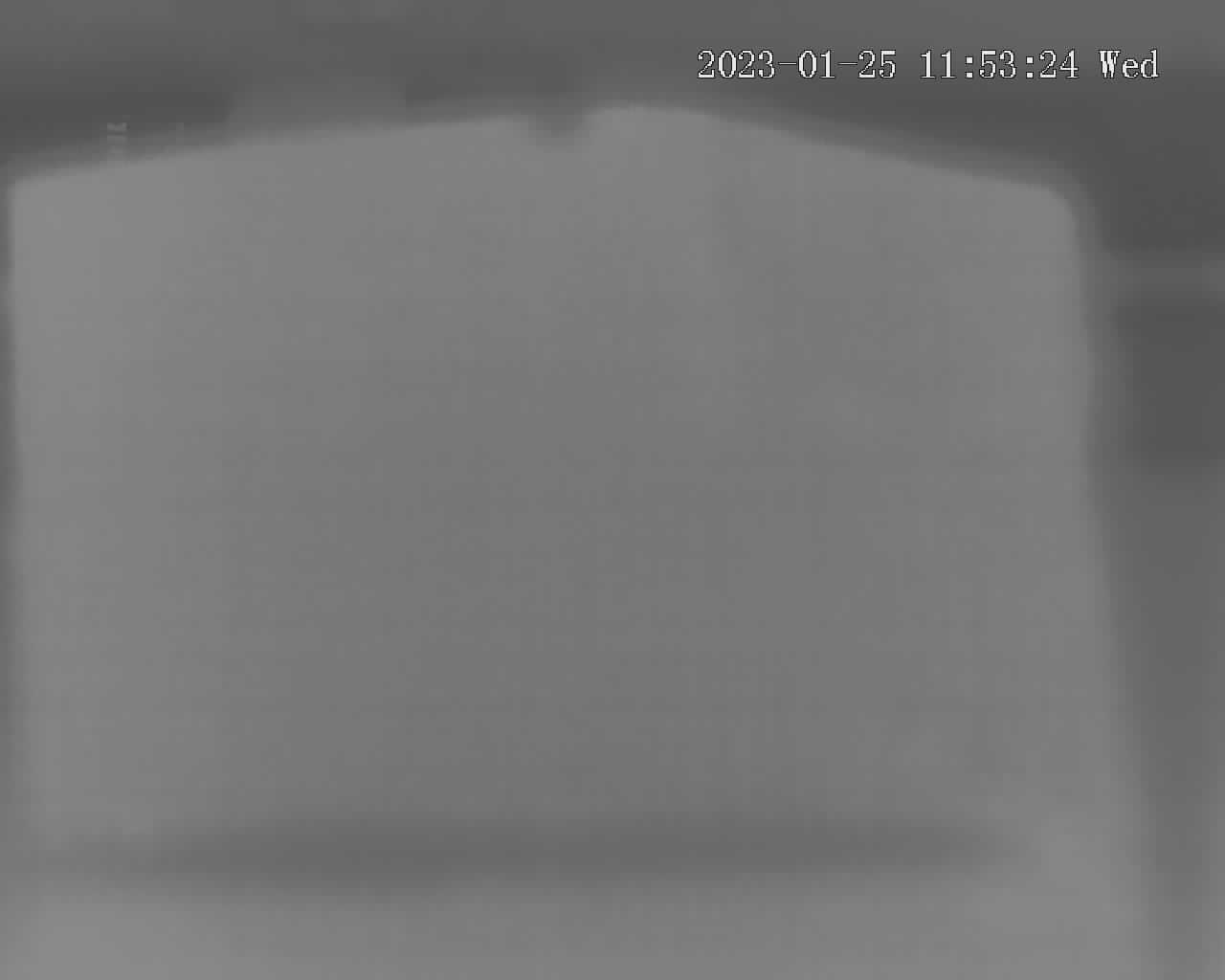}
        \includegraphics[scale=0.0847]{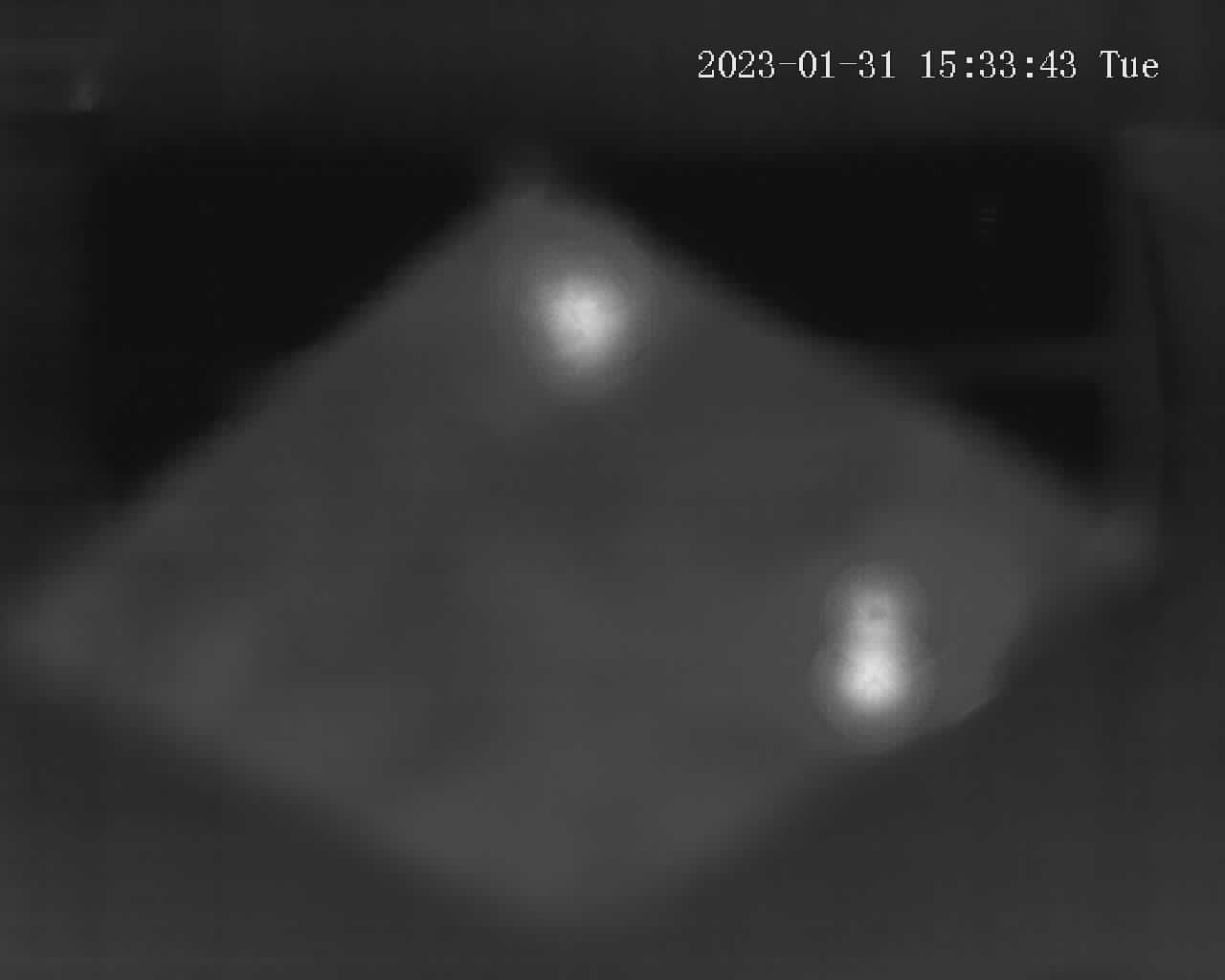}
        \includegraphics[scale=0.0847]{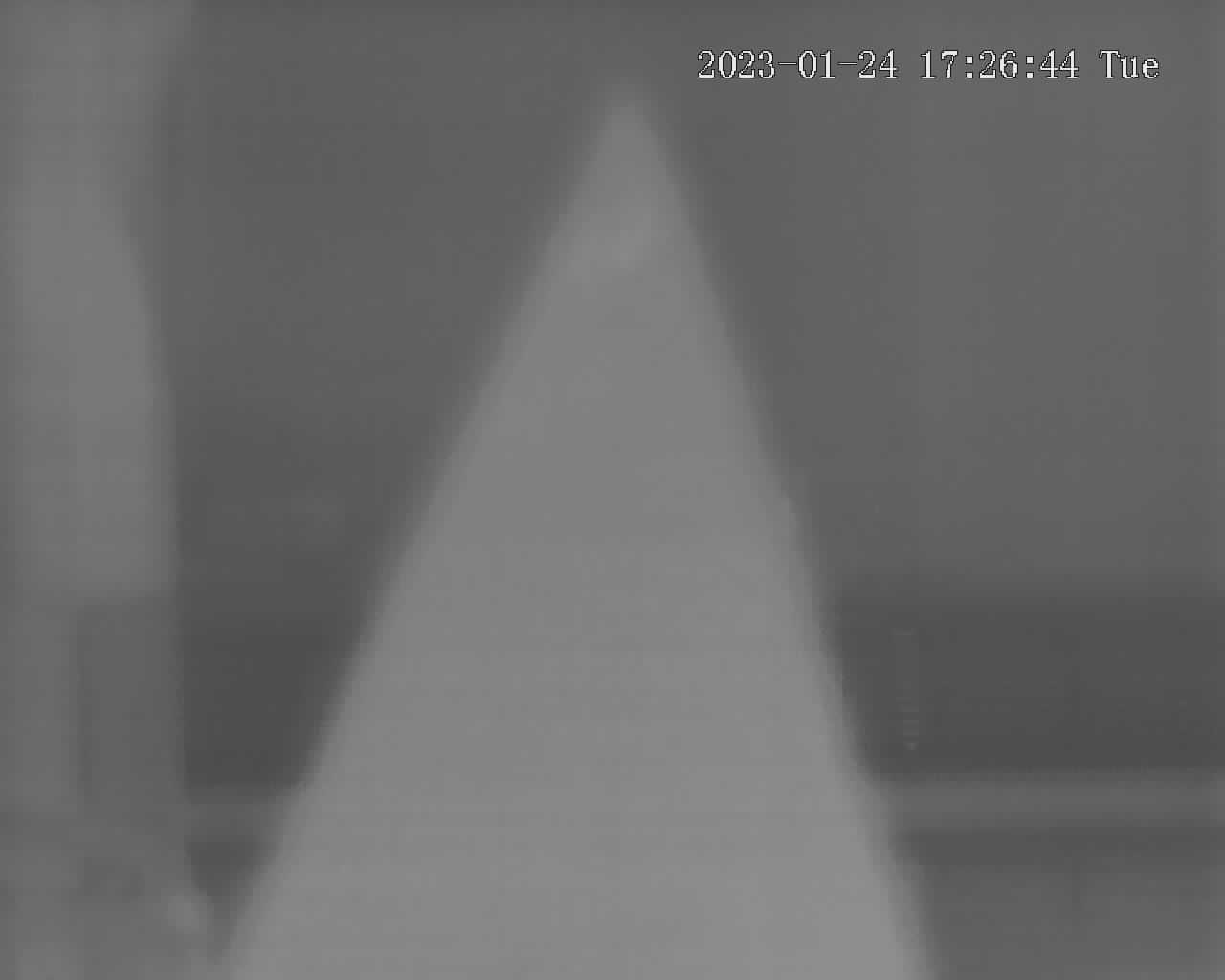}
        \includegraphics[scale=0.0847]{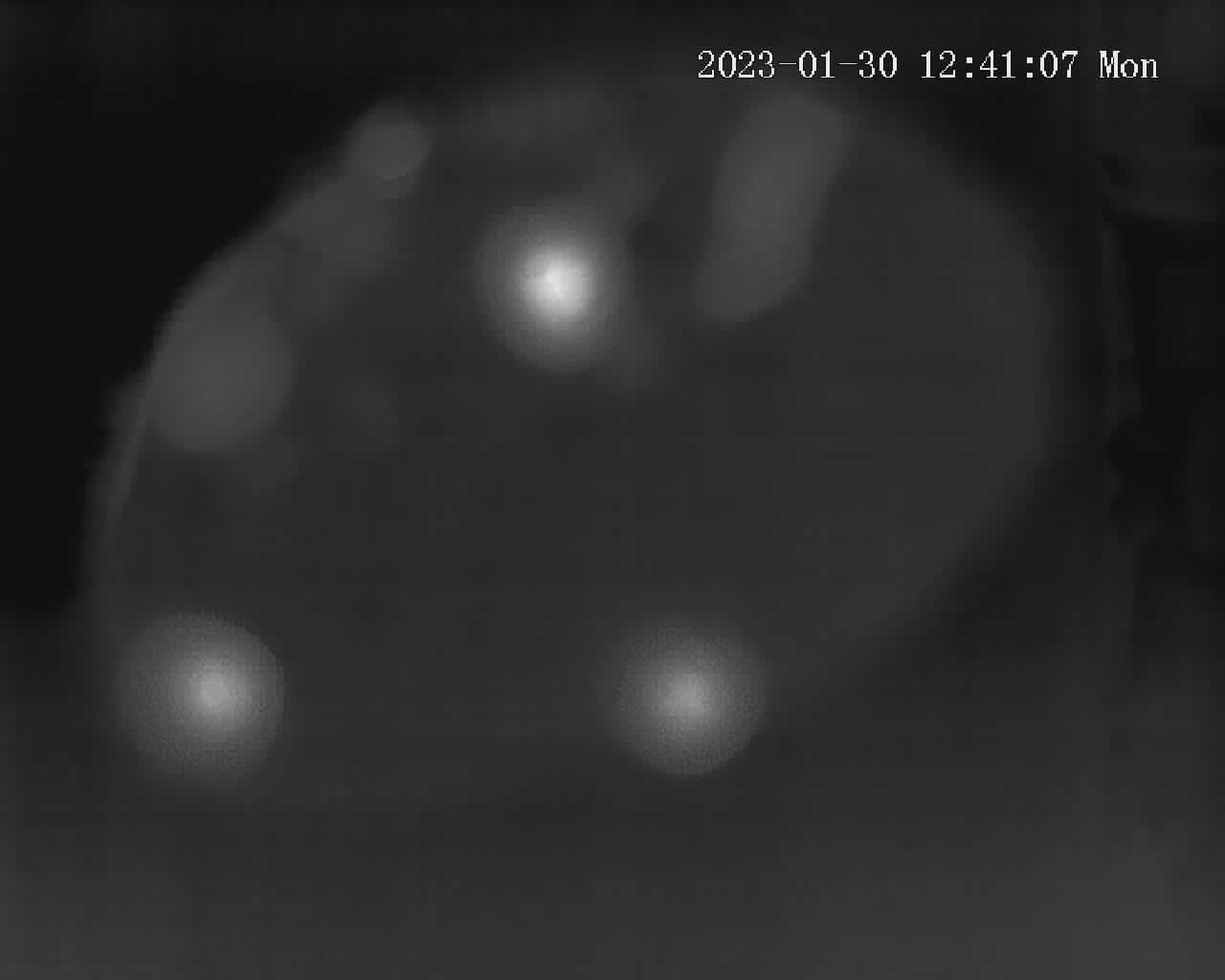}
        \caption{}
        \label{fig:ours_samples}
    \end{subfigure}
    \caption{(a) Sample images from the AAU VAP Trimodal Dataset, (b) Sample images from Thermal Images for Hotspot Detection (TIHD) dataset.}
    \label{fig:data_samples}
\end{figure}

The AAU VAP Trimodal Dataset \citep{palmero2016multi} consists of images of different room settings in which zero, one, two or three people are present and are interacting with each other or with the objects in the scene. This dataset consists of three modalities, namely, RGB images, thermal images, and depth images. Only the thermal modality was utilized for training our model. The images are from three scenes, with two scenes situated in a meeting room with minimum natural lighting, and one scene situated in a room with windows and lots of natural light. 
The purpose of creation of this dataset was to enable the detection and segmentation of people by combining details and information from the three modalities. The thermal images in this dataset were captured using an Axis Q1922 camera. The dimensions of each image is 640x480. The thermal modality consisted of 11540 images. 10000 unlabelled thermal images were used for training the model. From the same dataset, a different set of 1800 labelled images were used for fine-tuning the classifier network. These 1800 images consisted of augmented labelled images as well. Figure \ref{fig:proxy_samples} shows a few sample images from this dataset.

\subsubsection{Thermal Images for Hotspot Detection (TIHD) Dataset}
The  Thermal Images for Hotspot Detection (TIHD) data was created by us. This dataset contains 7361 images in total. Of these 4017 unlabelled images were used for training the model (encoder), and 3344 labelled images for fine-tuning. The dataset was created by capturing images of aluminium metal sheets of different shapes and sizes from an approximate distance of 2 meters. The shapes used were: rectangle, triangle, oval and rhombus. The dimensions of the subjects were as follows: Rectangle - 33cm x 55cm. Triangle - base of 38cm, and perpendicular height from centre of the base of 52cm. Rhombus - side length of 26cm. Oval - Major axis length of 38cm, minor axis length of 32cm.
In order to create hotspots, the metal sheets were heated at random regions on their surface, either once, twice, or thrice. The thermal camera used for creating this dataset was the InfiRay IRS-PT264 Dual Spectrum Thermal Camera. Each captured image is a heatmap of dimensions 1280x1024. Figure \ref{fig:ours_samples} shows a few sample images from this dataset. 

\subsection{Training the encoder and predictor}\label{trainingencandpred}
In one of the SimSiam frameworks the encoder and the predictor were trained by minimizing the cost function in (\ref{eq:finalloss}), whereas in the other one they were trained by minimizing the cost function in (\ref{eq:finalcosinesimilarity}). For the compound loss equation given in (\ref{eq:finalloss}), the best weight $\beta$ was determined empirically to be $\frac{2}{3}$. For this, estimate of the weight was conducted to investigate the effects of different loss function weights on the model performance. Three weight ratios were selected at intervals of approximately 0.3 to cover a wide range of configurations. Specifically, the weight ratios selected were $\frac{1}{3}, \frac{2}{3}, \frac{17}{18}$. Upon training three different models with these weights, and then testing the corresponding encoders on a classification task on the TIHD dataset, we got accuracies of 90.48\%, 92.86\%, and 90.48\% respectively. Therefore, a weight ($\beta$) of $\frac{2}{3}$ was picked.
A reasonable loss curve representing the training undertaken without the use of a large batch size, or any explicitly defined negative samples was obtained, which without the use of a stop-gradient would have degenerated rapidly. With this we were able to successfully train the encoder and also train the predictor. 
The encoders of these two trained SimSiam models would be combined into an ensemble network for the purpose of classification.

The batch size used for training was 81. The Stochastic Gradient Descent Optimizer was used, with a constant learning rate of 0.0008 for the AAU-VAP dataset, and 0.001 for the TIHD dataset. A momentum of 0.6 was also used. Training was conducted for 200 epochs. The duration of each iteration was approximately 65s. 

\subsection{Fine-tuning the classifier}
\begin{figure}
    \centering
    \begin{subfigure}{.245\textwidth}
        \centering
        \includegraphics[scale=0.243]{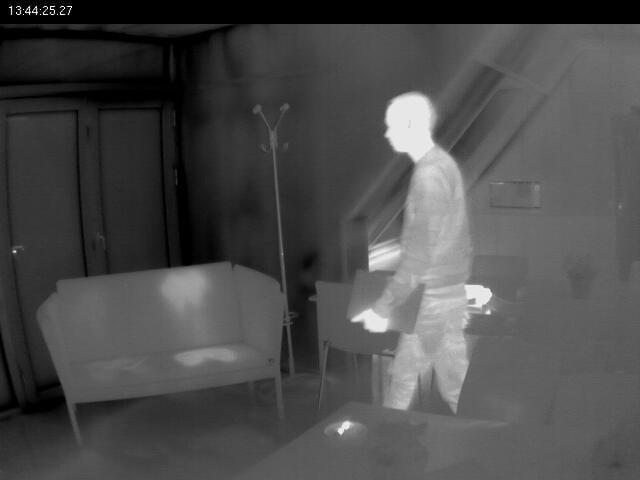}
        \caption{}
        \label{fig:normalDS}
    \end{subfigure}
    \begin{subfigure}{.245\textwidth}
        \centering
        \includegraphics[scale=0.243]{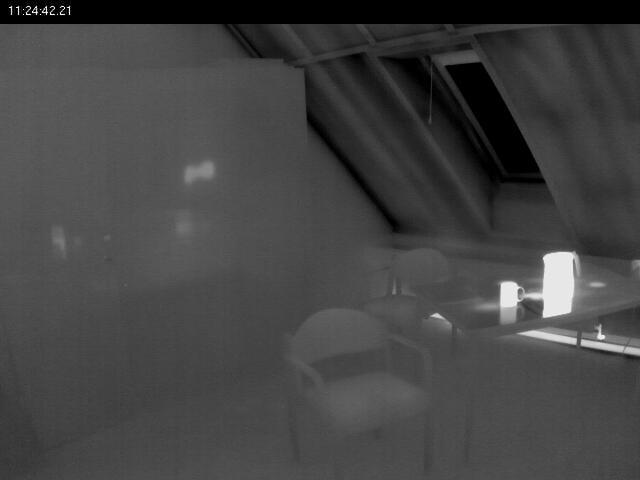}
        \caption{}
        \label{fig:normalDS2}
    \end{subfigure}
    \begin{subfigure}{.245\textwidth}
        \centering
        \includegraphics[scale=0.243]{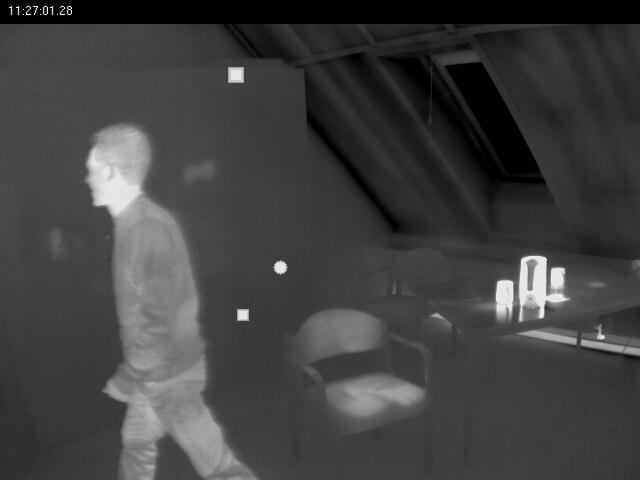}
        \caption{}
        \label{fig:anomalousDS}
    \end{subfigure}
    \begin{subfigure}{.245\textwidth}
        \centering
        \includegraphics[scale=0.243]{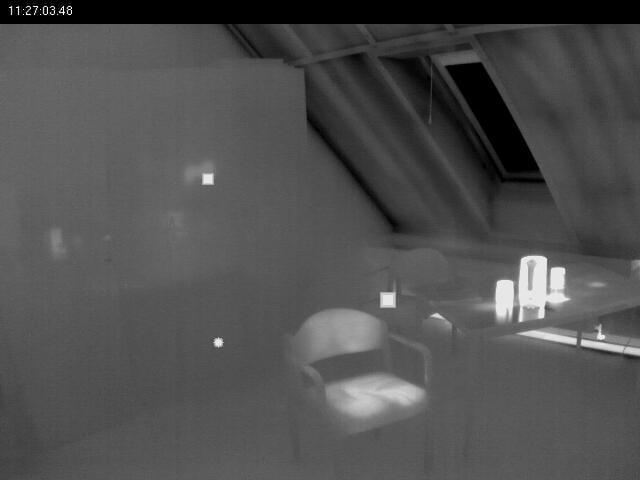}
        \caption{}
        \label{fig:anomalousDS2}
    \end{subfigure}
    \begin{subfigure}{.245\textwidth}
        \centering
        \includegraphics[scale=0.088]{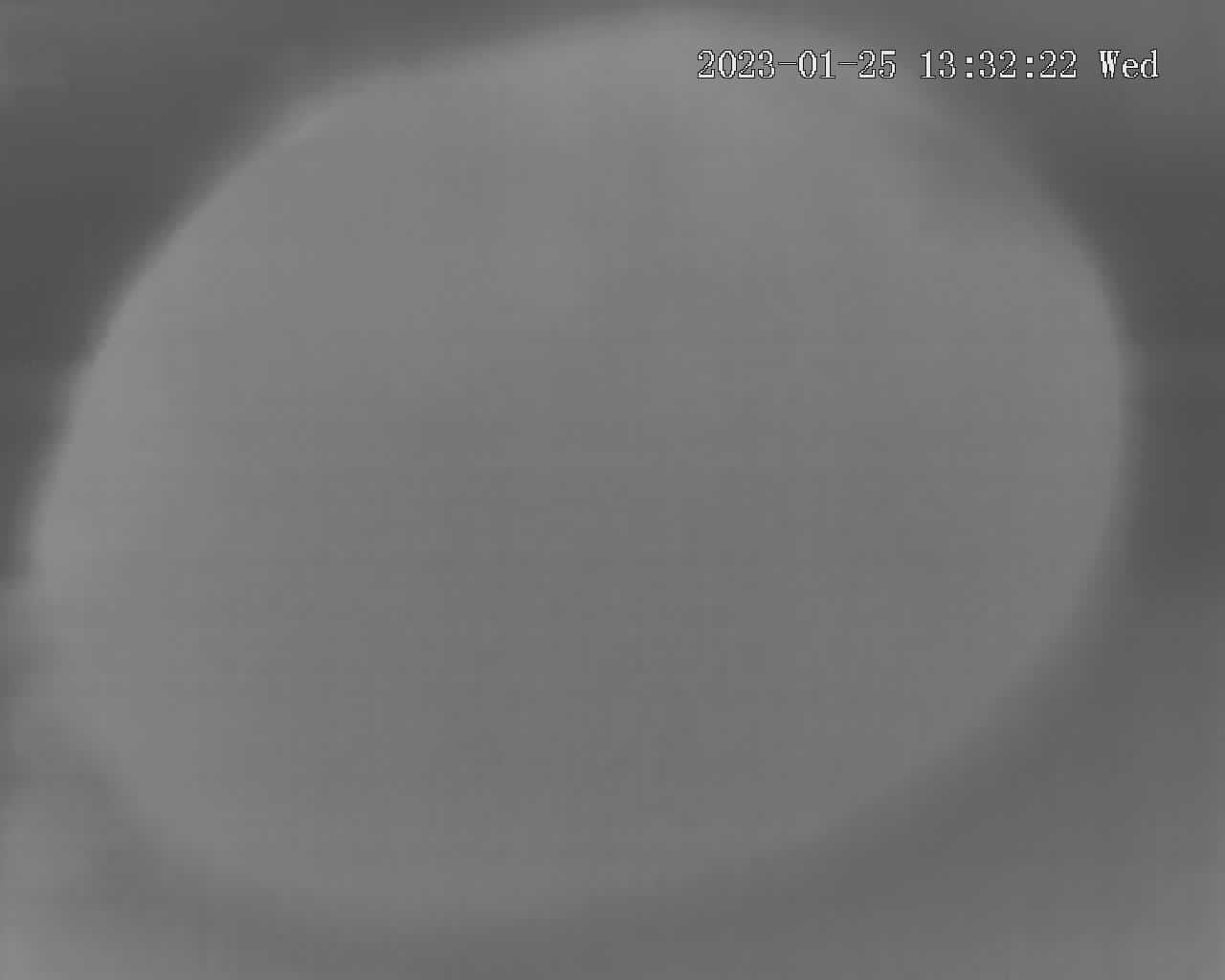}
        \caption{}
        \label{fig:normalourDS}
    \end{subfigure}
    \begin{subfigure}{.245\textwidth}
        \centering
        \includegraphics[scale=0.088]{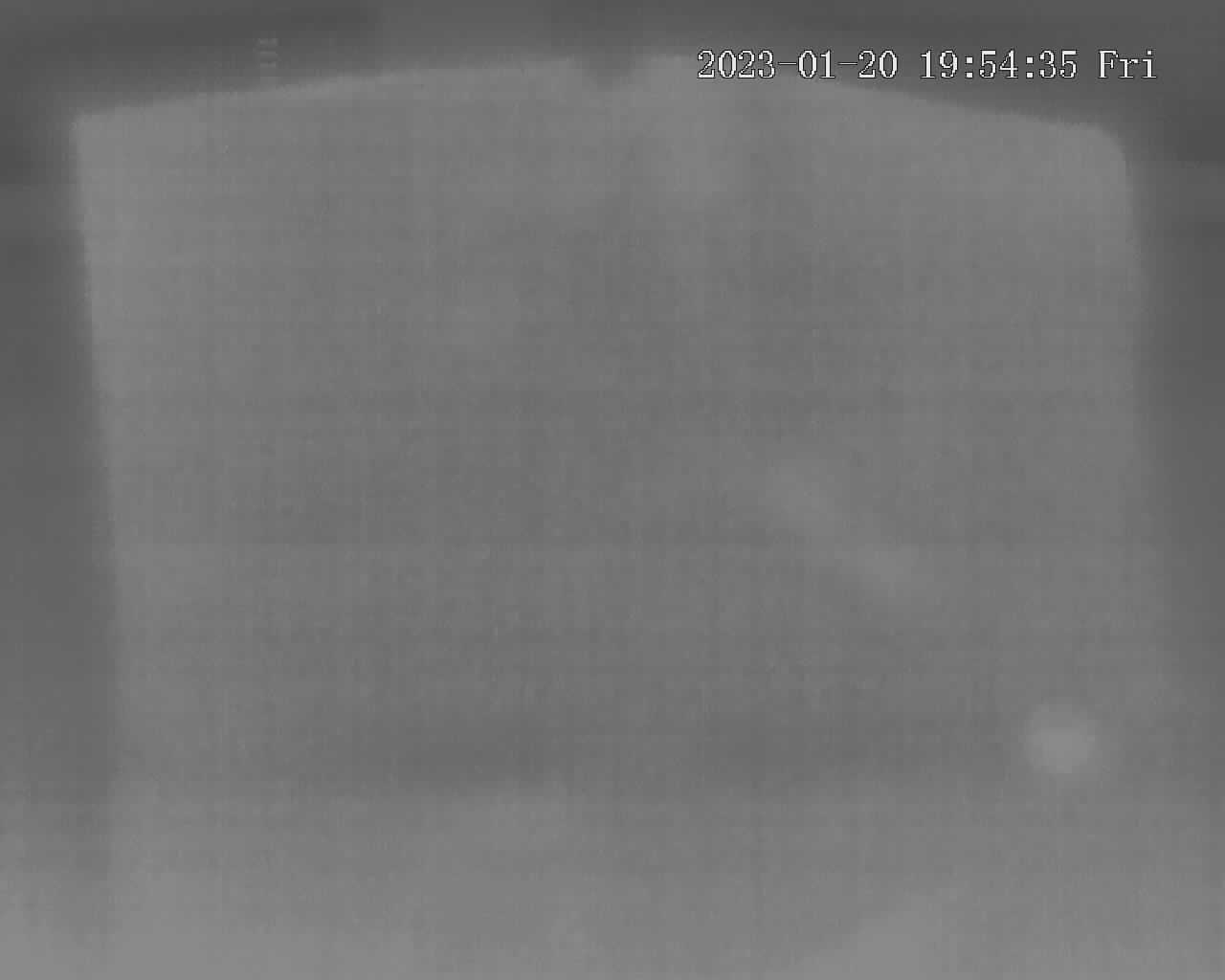}
        \caption{}
        \label{fig:normalourDS2}
    \end{subfigure}
    \begin{subfigure}{.245\textwidth}
        \centering
        \includegraphics[scale=0.088]{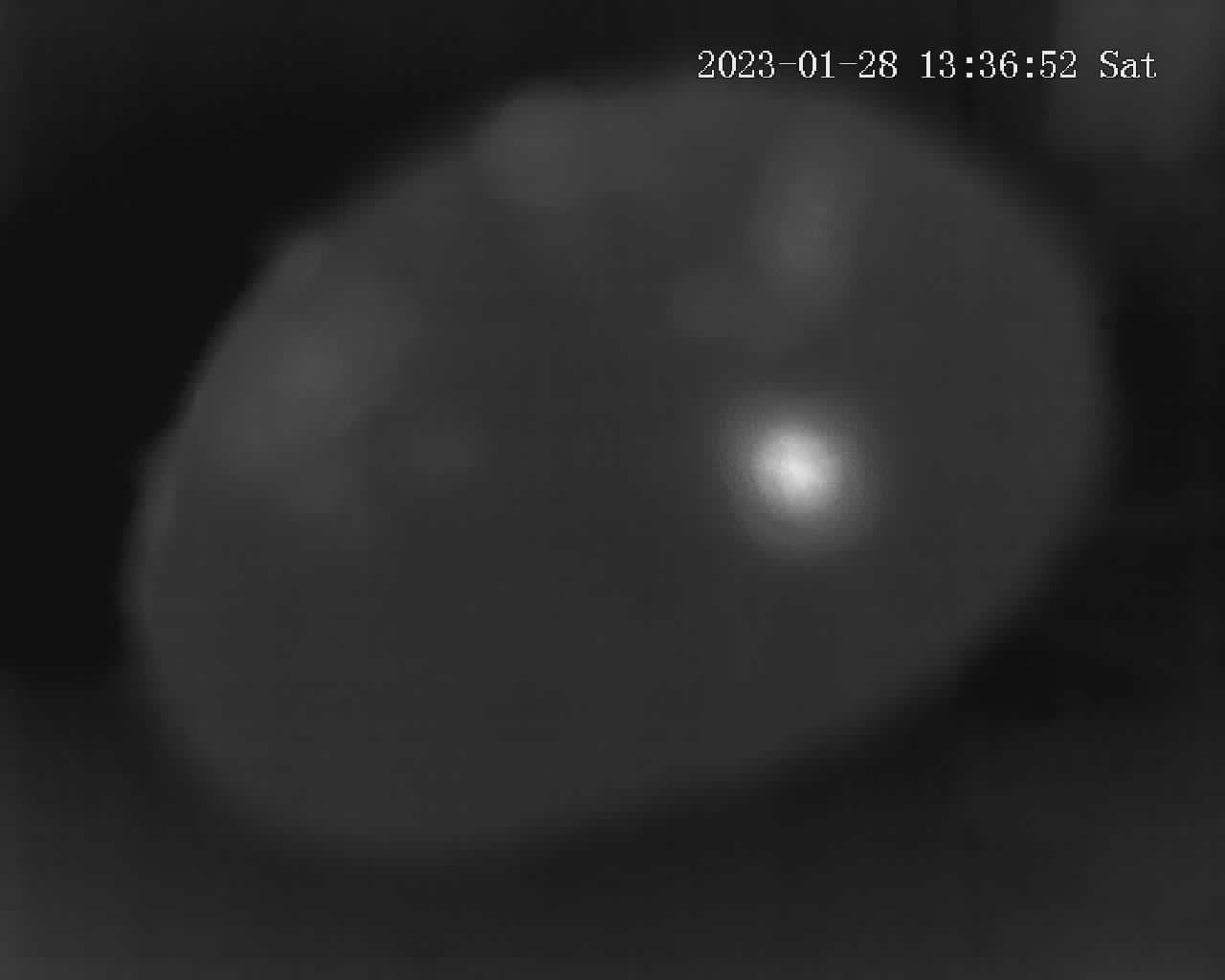}
        \caption{}
        \label{fig:hotspotourDS}
    \end{subfigure}
    \begin{subfigure}{.245\textwidth}
        \centering
        \includegraphics[scale=0.088]{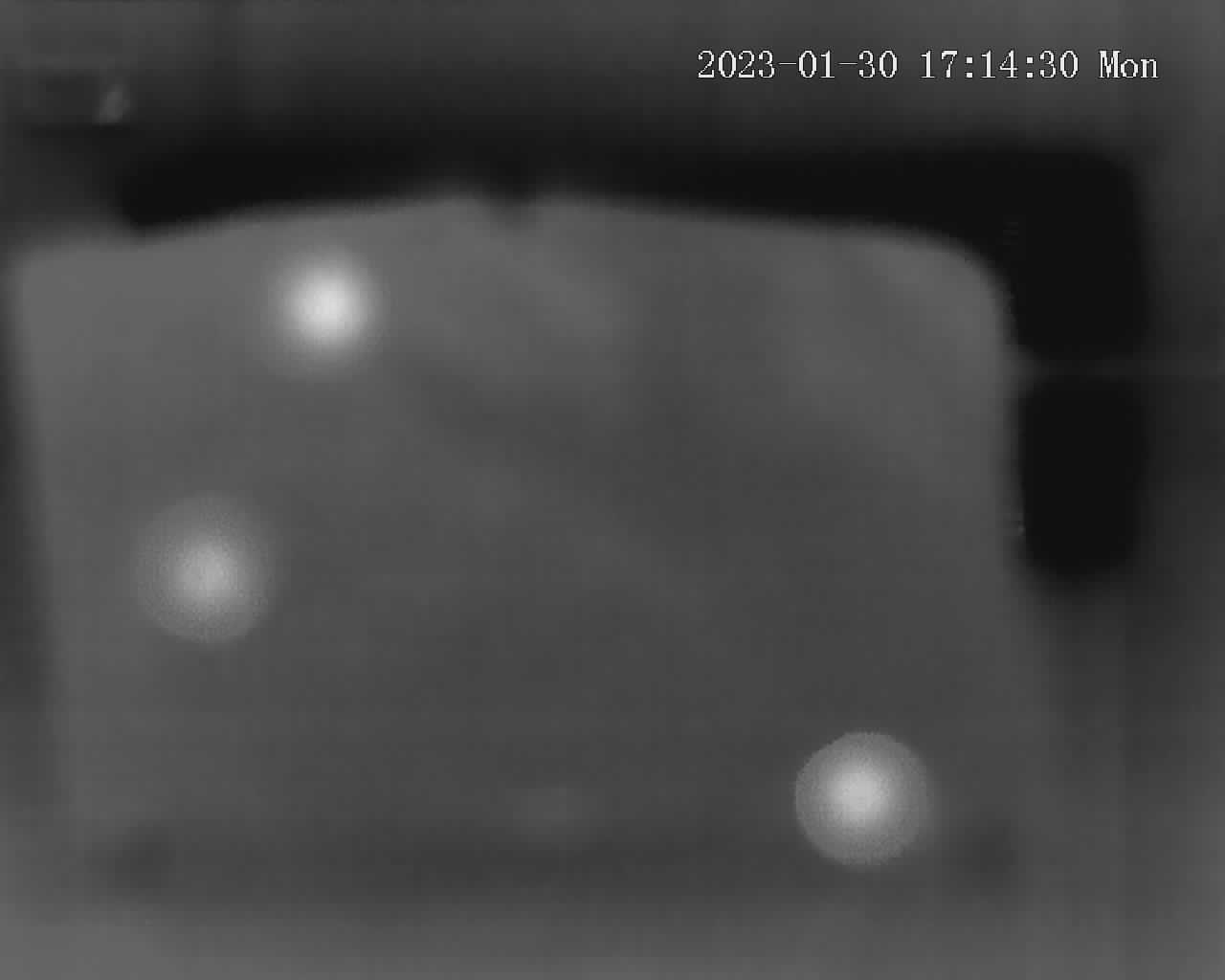}
        \caption{}
        \label{fig:hotspotourDS2}
    \end{subfigure}
    \caption{Examples of normal images (\ref{fig:normalDS}, \ref{fig:normalDS2}, \ref{fig:normalourDS}, \ref{fig:normalourDS2}) and images containing hotspots (\ref{fig:anomalousDS}, \ref{fig:anomalousDS2}, \ref{fig:hotspotourDS}, \ref{fig:hotspotourDS2}) from AAU-VAP dataset (Row 1), and Thermal Images for Hotspot Detection (TIHD) dataset (Row 2).}
    \label{fig:normalandanomalousDS}
\end{figure}

For fine-tuning the classifier, the 1800 images were labelled as anomalous or normal from the AAU VAP dataset. The fine-tuning task was defined as a 2-way softmax based classification task. Figures \ref{fig:normalDS} and \ref{fig:anomalousDS} shows an example of a normal image and an anomalous image (image containing hotspot) in this dataset. 

The encoders used for this downstream task were taken from the two trained modified SimSiam models and loaded with the trained encoder weights so that the learned representations from the training could be utilized by the classifier while fine-tuning on the new labelled dataset. The fine-tuning was conducted for 200 epochs as well along with early stopping constraints.
It was observed that the training and validation losses decreased reasonably with the number of epochs and the training and validation accuracies increased with the number of epochs up until a certain point. 

The average accuracy achieved through this method on the AAU-VAP dataset was around 94\%.
Similarly, training and fine-tuning was conducted for TIHD dataset. The average accuracy of fine-tuning on TIHD dataset was found to be around 97\%.
Sample images (normal and hotspot present) from the thermal dataset created by us are shown in Figures \ref{fig:normalourDS}, \ref{fig:normalourDS2}, \ref{fig:hotspotourDS}, and \ref{fig:hotspotourDS2}.

The training step is an important step for the encoder to be able to learn meaningful characteristics of the data. To test this, an XceptionNet based encoder was 'fine-tuned' without any pre-training. It was observed that the training degenerated fast and stops, with not much learning taking place in the model. This comparison shows the effectiveness and benefits of using this contrastive learning method with unlabelled data for this kind of classification tasks.

\subsection{Ablation Study}\label{ablation}
We conducted an ablation study of the developed method in order to evaluate the contributions of the different elements of the process and the modified SimSiam framework. For this, starting from the complete finished process, elements were removed sequentially till a single SimSiam framework with a ResNet50 backbone remained.

All in all, there were three SimSiam frameworks trained under different conditions for the purpose of experimentation. Of these, the encoders of two (Models 1 and 2) were used in different ensemble models for the ablation study. The list of these three models is as follows:
\begin{enumerate}
  \item Model 1: SimSiam with Xception based encoder, trained with compound loss.
  \item Model 2: SimSiam with Xception based encoder, trained with regular loss.
  \item Model 3: SimSiam trained with regular loss but with ResNet50 as the encoder.
\end{enumerate}
where the \textit{regular loss} is the cosine similarity loss in  (\ref{eq:finalcosinesimilarity}), and \textit{compound loss} represents a weighted sum of the cosine similarity loss and the cross-entropy loss in
(\ref{eq:finalloss}).

This ablation study consisted of the following:
\begin{enumerate}
  \item Our method (Ensemble of the encoders of \textit{Model 1} and \textit{Model 2})
  \item Removal of cross-entropy loss in the second SimSiam framework. (Ensemble of the encoders of two separately trained \textit{Model 2})
  \item Removal of the second SimSiam framework. (\textit{Model 2})
  \item Replacing XceptionNet with ResNet50 as the backbone in the remaining single SimSiam framework. (\textit{Model 3})
\end{enumerate}
The accuracy, precision, sensitivity, specificity and F-scores were calculated for each of the models in the ablation process as shown in (\ref{eq:eval_metrics}).
\begin{equation}\label{eq:eval_metrics}
\begin{split}
    &Accuracy = \frac{TP + TN}{TP + FP + TN + FN} \\
    &Precision = \frac{TP}{TP + FP} \\
    &Sensitivity = \frac{TP}{TP + FN} \\
    &Specificity = \frac{TN}{TN + FP} \\
    &F-Score = \frac{TP}{TP + \frac{1}{2}(FP + FN)}
\end{split}
\end{equation}
where TP, TN, FP, and FN refer to true positives, true negatives, false positives, and false negatives respectively.
Table \ref{table:ablation_tihd} summarizes the results of the ablation study on the TIHD dataset, whereas Table \ref{table:ablation_aauvap} summarized the results on the AAU-VAP dataset.

\begin{table}
    \centering
    \caption{Ablation study result metrics on the TIHD dataset. First row shows our method and subsequent rows show metrics after removal of the cross-entropy loss, the ensemble network, and the XceptionNet backbone respectively.}
    \begin{tabular}{cccccc}
        \hline
         & Accuracy & Precision & Sensitivity & Specificity & F-Score\\ [0.5ex]
        \hline\hline
        Our method & 0.97 & 1.00 & 0.94 & 1.00 & 0.97\\
        \hline
        without Cross-entropy loss & 0.91 & 0.95 & 0.86 & 0.96 & 0.90\\
        \hline
        without the Ensemble & 0.90 & 0.85 & 0.96 & 0.84 & 0.90\\
        \hline
        without XceptionNet & 0.89 & 0.91 & 0.86 & 0.92 & 0.88\\
        \hline
    \end{tabular}
    
    \label{table:ablation_tihd}
\end{table}
\begin{table}
    \centering
     \caption{Ablation study result metrics on the AAU-VAP dataset. First row shows our method and subsequent rows show metrics after removal of the cross-entropy loss, the ensemble network, and the XceptionNet backbone respectively.}
    \begin{tabular}{cccccc}
        \hline
         & Accuracy & Precision & Sensitivity & Specificity & F-Score\\ [0.5ex]
        \hline\hline
        Our method & 0.94 & 0.92 & 0.96 & 0.92 & 0.94\\
        \hline
        without Cross-entropy loss & 0.90 & 0.88 & 0.92 & 0.88 & 0.90\\
        \hline
        without the ensemble & 0.90 & 0.87 & 0.94 & 0.86 & 0.90\\
        \hline
        without XceptionNet & 0.85 & 0.92 & 0.76 & 0.94 & 0.83\\
        \hline
    \end{tabular}
    \label{table:ablation_aauvap}
\end{table}
The weights of the encoder predictions in the ensemble models from Table \ref{table:ablation_tihd} and Table \ref{table:ablation_aauvap} were found using grid search for maximizing the accuracy.

From the metrics in Table \ref{table:ablation_tihd} and \ref{table:ablation_aauvap} it can be seen that our proposed method which involves an ensemble network and a compound loss consisting of the negative cosine similarity and the cross-entropy loss performs the best. Once the different modules were removed one by one, the accuracy and the F-score seem to decrease. It can also be seen that using XceptionNet as the encoder leads to better performance than using ResNet50. Specifically, the F-score increases from 0.88 to 0.90 for the TIHD dataset, and from 0.83 to 0.90 for the AAU-VAP dataset. Furthermore, once the cross-entropy loss is added to the loss function, it is observed that the accuracy shows significant improvement on both datasets. The accuracies increase from 91\% to 97\%, and from 90\% to 94\% on the TIHD and AAU-VAP datasets respectively. Therefore, major improvement was seen because of the inclusion of the compound weighted loss function.

\subsection{Comparison of hotspot detection with existing supervised learning techniques}
For the purpose of comparison with existing supervised learning techniques, machine learning models like SVM (support vector machines), Logistic Regression, and Random Forest were chosen, along with CNN models like VGG16 \citep{simonyan2014very}, XceptionNet, EfficientNet \citep{tan2019efficientnet}, ResNet \citep{he2016deep} and a custom neural network. Given that it is a classification task, the AUC (Area under the curve) - ROC (Receiver operating characteristics) plot was plotted for each model, and the AUC Score was compared alongside other metrics of importance. 

All the models were tested on a test dataset consisting of 100 images from the two datasets respectively. Based on the results on the testing dataset, model accuracy, precision, sensitivity, specificity and F-Score were calculated. This was done by first counting the number of true positives (TP), true negatives (TN), false positives (FP) and false negatives (FN). Following this, the metrics were calculated.

At first, the created model was compared to the existing supervised machine learning and neural network models trained on the AAU VAP Thermal Dataset. 
Upon comparison with the existing machine learning models, it was found that our method outperforms all of the chosen machine learning techniques on all the metrics when it comes to the AAU VAP Dataset. Table \ref{table:Comparison_ml_nn_aau} summarizes the comparison of our model with the existing supervised machine learning techniques and the existing neural networks on the AAU VAP dataset.  
\begin{table}
\centering
\caption{Comparison of performance of our model with existing supervised machine learning techniques and neural network models on the AAU-VAP Thermal Dataset.}
\begin{tabular}{c  c  c  c  c  c  c} 
 \hline
  & Accuracy & Precision & Sensitivity & Specificity & F-Score & AUC Scores\\ [0.5ex] 
 \hline\hline
 SVM & 0.72 & 0.64 & \textbf{1.00} & 0.44 & 0.78 & 0.73\\ 
 \hline
 Rand. Forest & 0.50 & 0.50 & 0.38 & 0.62 & 0.43 & 0.50\\
 \hline
 Log. Regression & 0.58 & 0.65 & 0.34 & 0.82 &  0.44 & 0.58\\
 \hline\hline
 VGG16 & 0.85 & 0.87 & 0.82 & 0.88 & 0.84 & 0.91\\
\hline
XceptionNet & 0.50 & 0.50 & 1.00 & 0.00 & 0.66 & 0.50\\
\hline
EfficientNet & \textbf{0.94} & \textbf{1.00} & 0.88 & \textbf{1.00} & 0.93 & \textbf{0.96}\\
\hline
ResNet & 0.93 & \textbf{1.00} & 0.86 & \textbf{1.00} & 0.92 & \textbf{0.97}\\ 
\hline
Custom NN & 0.88 & 0.95 & 0.80 & 0.96 & 0.86 & 0.91\\
\hline
Ours & \textbf{0.94} & 0.92 & \textbf{0.96} & 0.92 & \textbf{0.94} & \textbf{0.94}\\
\hline
\end{tabular}
\label{table:Comparison_ml_nn_aau}
\end{table}

\begin{figure}
    \centering
    \begin{subfigure}{0.46\textwidth}
        \includegraphics[scale=0.465]{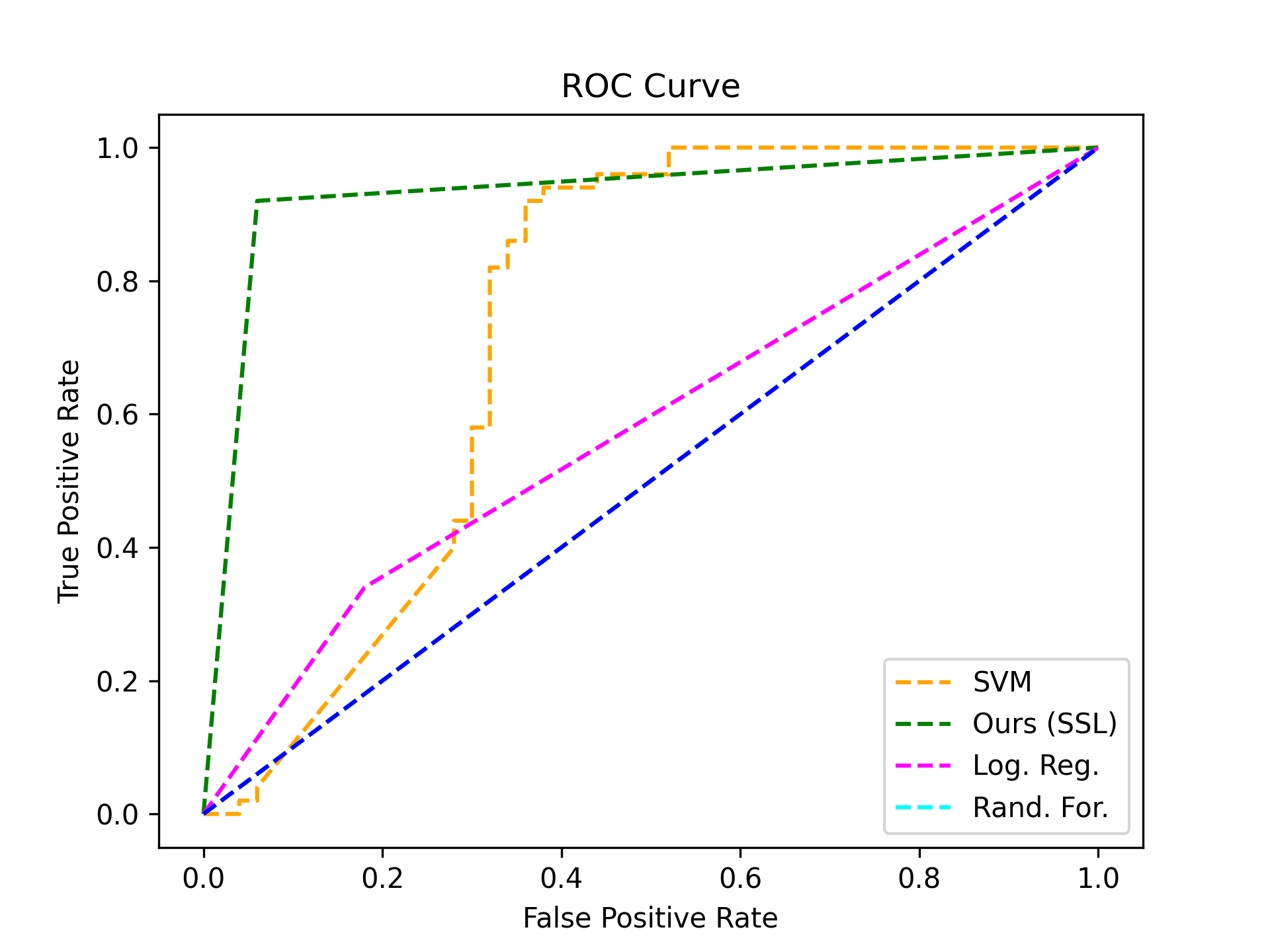}
        \caption{}
        \label{fig:AUC-ROC_ml_AAU}
    \end{subfigure}
    \hspace{0.01in}
    \begin{subfigure}{0.46\textwidth}
        \includegraphics[scale=0.465]{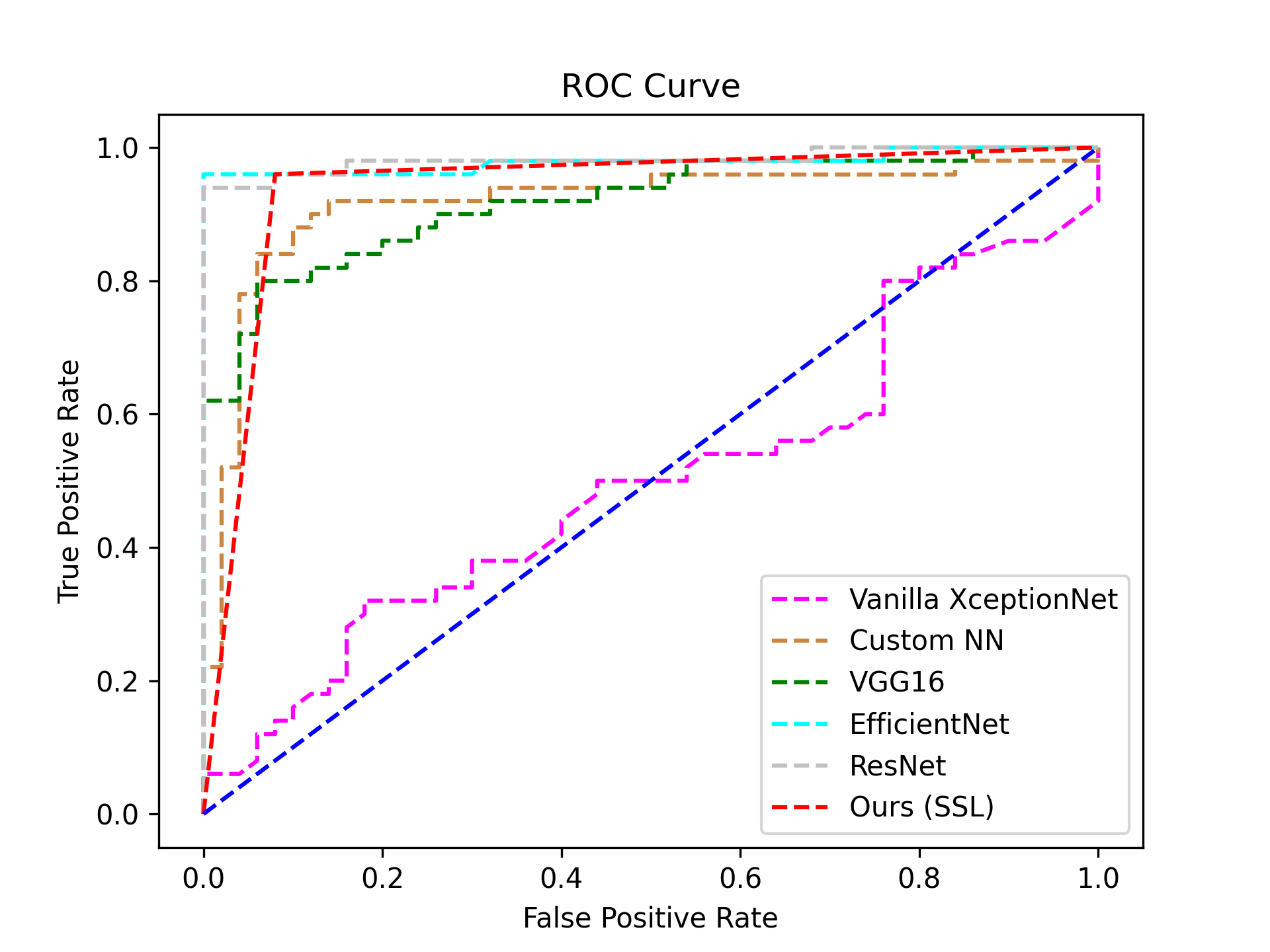}
        \caption{}
        \label{fig:AUC-ROC_nn_AAU}
    \end{subfigure}\\
    \caption{AUC-ROC plots of different (a) supervised machine learning and (b) neural network models trained on the AAU-VAP thermal dataset for comparison with our self-supervised learning model.}
    \label{fig:auc-roc-aau}
\end{figure}

Upon plotting the AUC-ROC curve from the results obtained by aforementioned machine learning models as well as by our model, the plot shown in Figure \ref{fig:AUC-ROC_ml_AAU} was obtained. The corresponding AUC (area under the curve) scores of each model are also mentioned in Table \ref{table:Comparison_ml_nn_aau}. From the AUC-ROC plot in Figure \ref{fig:AUC-ROC_ml_AAU}, as well as from the AUC scores, it can be seen that the area under the curve that represents our model (green) is much larger than the area under the curve of each of the other models. The SVM curve (yellow) shows reasonable class separation capability whereas the Logistic Regression curve shows little class separation capability in this case. The curve for the Random Forest model is hidden behind the reference blue curve (representing equal false positive and true positive rate) which means that it has almost no class separation capability. This confirms that our model outperforms the existing supervised machine learning techniques for classification.
Secondly, in a similar manner a comparison with existing supervised convolution neural network models was also conducted. Upon comparison with these models, it was found that our model performs in a competitive manner compared to them. Our model is able to identify a greater number of true positives compared to these models because of the training, and is able to perform competitively in the other metrics as well. Table \ref{table:Comparison_ml_nn_aau} summarizes the results of this comparison.

As seen from Table \ref{table:Comparison_ml_nn_aau}, the performance of our model is comparable to the performance of EfficientNet as is shown by the F-scores of the two models (0.94 and 0.93 respectively). The accuracy and sensitivity of our model is the highest among the different models, except for EfficientNet whose accuracy is the same as our model's. The custom neural network which consists of 3 sets of 2D convolution and max pooling layers, followed by a dropout layer, and ending with 2 dense layers also performs reasonably, and is close to the other models. This experiment proves that self-supervised learning is a competitive approach with respect to the supervised learning approaches when it comes to classification tasks. As far as XceptionNet is concerned, it is unable to learn any meaningful data characteristics for this dataset.

Upon plotting the AUC-ROC plot for this comparison, the plot shown in Figure \ref{fig:AUC-ROC_nn_AAU} is obtained. The corresponding AUC scores of each model are also mentioned in Table \ref{table:Comparison_ml_nn_aau}. From the AUC-ROC plot in Figure \ref{fig:AUC-ROC_nn_AAU}, as well as from the AUC scores, it can be seen that the area under the curve that represents our model (red), is highly competitive to the area under the curve of ResNet and EfficientNet (silver and cyan), which have slightly more AUC. The VGG16 AUC-ROC curve, and the custom NN curve show good class-separation capability as well. However, at certain points our model is able to provide a greater true positive rate coupled with a low false positive rate, because of the training, which increases its overall AUC score. The curve for the XceptionNet model is unable to rise to a reasonable level, and has minimal class-separation capability. This confirms that our model provides competitive performance when it comes to existing supervised neural network techniques. 

To bolster the efficacy of the self-supervised model, it was also compared with the mentioned supervised learning models by training on TIHD dataset. In order to make the classification task more challenging for the models, hotspot-like features of random shape and size were introduced in random regions of certain images of the dataset, and these images were labelled as 0 (do not contain hotspots). The models were then trained to identify the images with the real hotspots. The augmentations include introduction of salt and pepper noise and brightness changes apart from the introduction of fake anomalies in the images that do not contain a hotspot originally. 

\begin{figure}
    \centering
    \begin{subfigure}{0.46\textwidth}
        \includegraphics[scale=.465]{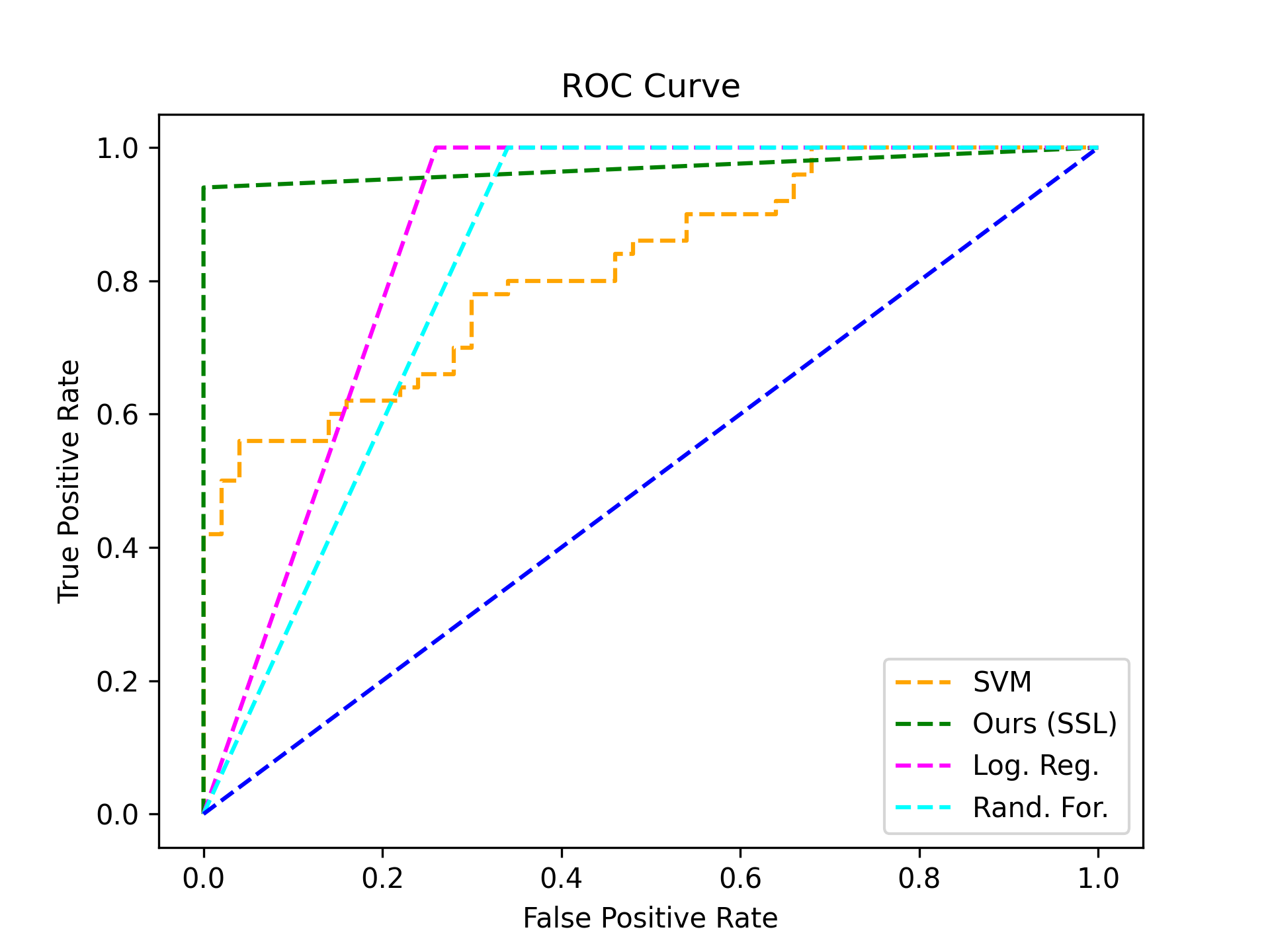}
        \caption{}
        \label{fig:AUC-ROC_ml_ours}
    \end{subfigure}
    \hspace{0.01in}
    \begin{subfigure}{0.46\textwidth}
        \includegraphics[scale=0.465]{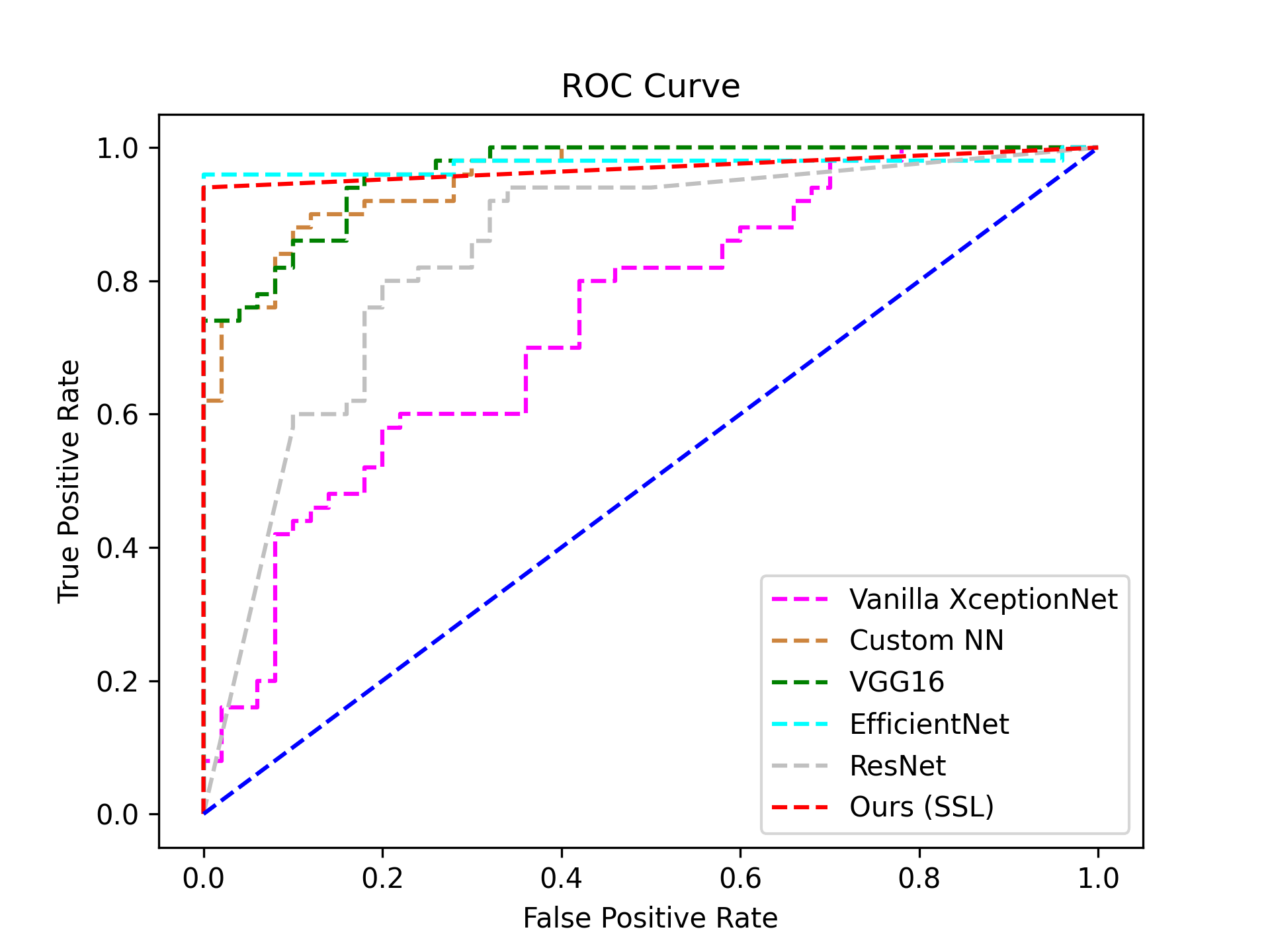}
        \caption{}
        \label{fig:AUC-ROC_nn_ours}
    \end{subfigure}\\
    \caption{AUC-ROC plots of different (a) supervised machine learning and (b) neural network models trained on TIHD thermal dataset for comparison with our self-supervised learning model.}
    \label{fig:auc-roc-ours}
\end{figure}

Comparison results with existing supervised machine learning models and neural networks are summarized in Table \ref{table:Comparison_ml_nn_ours}. It can be observed from this table that our model achieved the highest F-score of 0.97. Even though the machine learning models achieve a perfect sensitivity, i.e., they are able to identify all positives correctly, there performance with the negative samples is not the best. This impacts their overall performance.  Our model provides a reasonably balanced result when it comes to the positive and the negative samples and provide an overall dominant result. The logistic regression and the random forest models perform reasonably well with this dataset compared to the AAU-VAP dataset. In Table \ref{table:Comparison_ml_nn_ours}, the AUC scores have been summarized for the machine learning models trained on our dataset. Upon comparison with the traditional machine learning models, it can be observed that our model achieves the highest AUC score of 0.97, followed by logistic regression and random forest models. The same result has been visualized in Figure \ref{fig:AUC-ROC_ml_ours}, where it can be observed that our model (green) has the largest area under the curve.
\begin{table}
    \centering
    \caption{Comparison of performances of our model with existing supervised machine learning models and neural networks on the TIHD dataset}
    \begin{tabular}{ccccccc}
        \hline
        & Accuracy & Precision & Sensitivity & Specificity & F-Score & AUC Scores\\ [0.5ex]
        \hline\hline
        SVM & 0.73 & 0.92 & 0.50 & 0.96 & 0.65 & 0.82\\
        \hline
        Rand. Forest & 0.83 & 0.74 & \textbf{1.00} & 0.66 & 0.85 & 0.83\\
        \hline
        Log. Regression & 0.87 & 0.79 & \textbf{1.00} & 0.74 & 0.88 & 0.87\\
        \hline\hline
        VGG16 & 0.86 & 0.84 & 0.88 & 0.84 & 0.86 & 0.96\\
        \hline
        XceptionNet & 0.50 & 0.00 & 0.00 & 1.00 & 0.00 & 0.74\\
        \hline
        EfficientNet & 0.94 & \textbf{1.00} & 0.88 & \textbf{1.00} & 0.94 & \textbf{0.97}\\
        \hline
        ResNet & 0.90 & 0.90 & 0.90 & 0.90 & 0.90 & 0.90\\
        \hline
        Custom NN & 0.86 & 0.82 & 0.92 & 0.80 & 0.86 & 0.95\\
        \hline
        Ours & \textbf{0.97} & \textbf{1.00} & \textbf{0.94} & \textbf{1.00} & \textbf{0.97} & \textbf{0.97}\\
        \hline
    \end{tabular}
    \label{table:Comparison_ml_nn_ours}
\end{table}

The model was also compared to the existing supervised neural network models trained on the TIHD dataset.
It can be observed from Table \ref{table:Comparison_ml_nn_ours} that our model gives a highly competitive performance to the supervised neural network models like VGG16, EfficientNet, and the custom NN. Our model is able to achieve an F-score of 0.97, whereas EfficientNet achieves an F-score of 0.94. It was noted in TIHD dataset specifically, that the particularly challenging images that were misclassified by the existing supervised learning networks, were classified correctly by our self-supervised network. This point often led to an overall improvement in its performance compared to the other models. The AUC scores obtained for this comparison, and the corresponding AUC-ROC plot can be seen in Table \ref{table:Comparison_ml_nn_ours} and Figure \ref{fig:AUC-ROC_nn_ours} respectively. Again, from Figure \ref{fig:AUC-ROC_nn_ours} it can be observed that our model (red) has more points in regions with greater true positive rates and low false positive rates, leading to its better performance compared to the other models. Our model's performance is almost at par with the performance of EfficientNet.

With all these results, it can be seen that the self-supervised learning model so developed is suitable for use in an industrial setting such as in the oil and gas exploration industry where hotspot detection in metallic equipment is of immense importance. What works in favour of this method is its higher sensitivity and specificity compared to other models since it is of utmost importance to not miss out on any possible defects (few false negatives), as well as to not classify negatives as positives.

\subsection{Hotspot Isolation}
\begin{figure}
    \centering
    \begin{subfigure}{0.30\textwidth}
        \includegraphics[scale = 0.11]{hm13.jpg}
        \includegraphics[scale = 0.11]{hm15.jpg}
        \caption{}
        \label{fig:Grad-CAM col 1}
    \end{subfigure}
    \begin{subfigure}{0.30\textwidth}
        \includegraphics[scale = 0.11]{gradcamres_hm13.jpg}
        \includegraphics[scale = 0.11]{gradcamres_hm15.jpg}
        \caption{}
        \label{fig:Grad-CAM col 2}
    \end{subfigure}
    \begin{subfigure}{0.30\textwidth}
        \includegraphics[scale = 0.423]{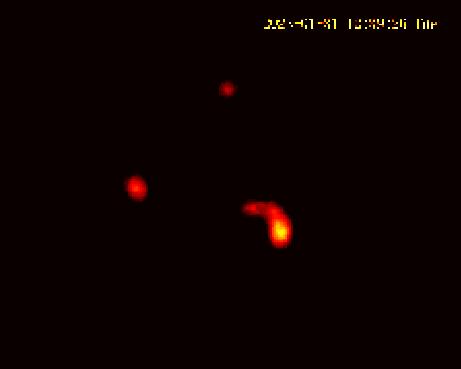}
        \includegraphics[scale = 0.423]{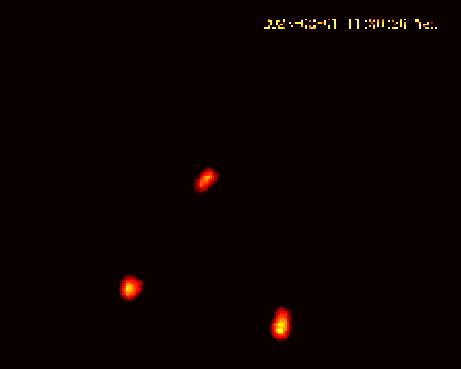}
        \caption{}
        \label{fig:Grad-CAM col 3}
    \end{subfigure}\\
    \caption{Examples of hotspot identification results. (a) the input images, (b) the attention heatmaps, (c) identified hotspots}
    \label{fig:Grad-CAM fig}
\end{figure}

Even though neural networks achieve reasonable performance, it is often unclear as to why a particular model makes particular decisions when provided with an input. In other words, neural networks can be thought of as 'black-box' models whose internal decision-making is often unknown. This is the reason why we have made use of the GradCAM \citep{selvaraju2017grad} model to see what exactly our model focuses on when provided with an input, to produce its classification result. The GradCAM model is able to output heatmaps showing areas of high attention. In our case, if the model performs as expected, the same heatmaps can also be used for pinpoint location detection of the hotspot in the provided image. Figure \ref{fig:Grad-CAM fig} shows the input images, and the isoloated hotspots from GradCAM outputs.

From Figure \ref{fig:Grad-CAM fig}, it can be observed in rows 1 and 2 that GradCAM is able to focus solely on the hotspots on the metal sheets for making its classification decision. Highest attention levels are depicted in red, and it can be seen that the hotspot centres are the regions in highest focus. These images also depict that different degrees of attention is placed on different degrees of hotness, with the highest attention being placed on the hottest regions. These results show that our model is able to make classification decisions by focussing on hotspots in a reasonable manner, and therefore works as expected. Its behaviour behind the scenes is not random or ambiguous. Furthermore, apart from model explainability, columns 2 and 3 of Figure \ref{fig:Grad-CAM fig} provide us with exact locations of the hotspots in the thermal images. Knowing the hotspot positions would facilitate any remedial work or further inspection that needs to be carried out in response, by providing engineers with a reference location of where the hotspot could be present.

\subsection{Comparison with existing hotspot isolation techniques}
Existing hotspot isolation techniques were reproduced for the purpose of comparison with our technique.
On comparison with the approach proposed in \cite{mohd2017application}, we found that our model is able to provide a more precise view of the hotspots present in a thermal image. The K-Means segmentation technique proposed by Mohd et al. is able to isolate the region containing hotspots as well, however the result still contains a lot of the background region (noise). The output of the technique can be observed in Figure \ref{fig:ComparisonsKMeans}. Therefore, after the use of the K-Means approach, further processing of the output will be required by the user to accurately identify the hotspots present in the image. Moreover, this method was not able to identify a few hotspots which got separated with the background, leaving holes in the foreground. On the other hand, the output of our method in \ref{fig:ComparisonsOurs} shows the precise isolation of all 3 hotspots.

\begin{figure}[!b]
    \centering
    \begin{subfigure}{\textwidth}
        \centering
        \includegraphics[scale=0.0733]{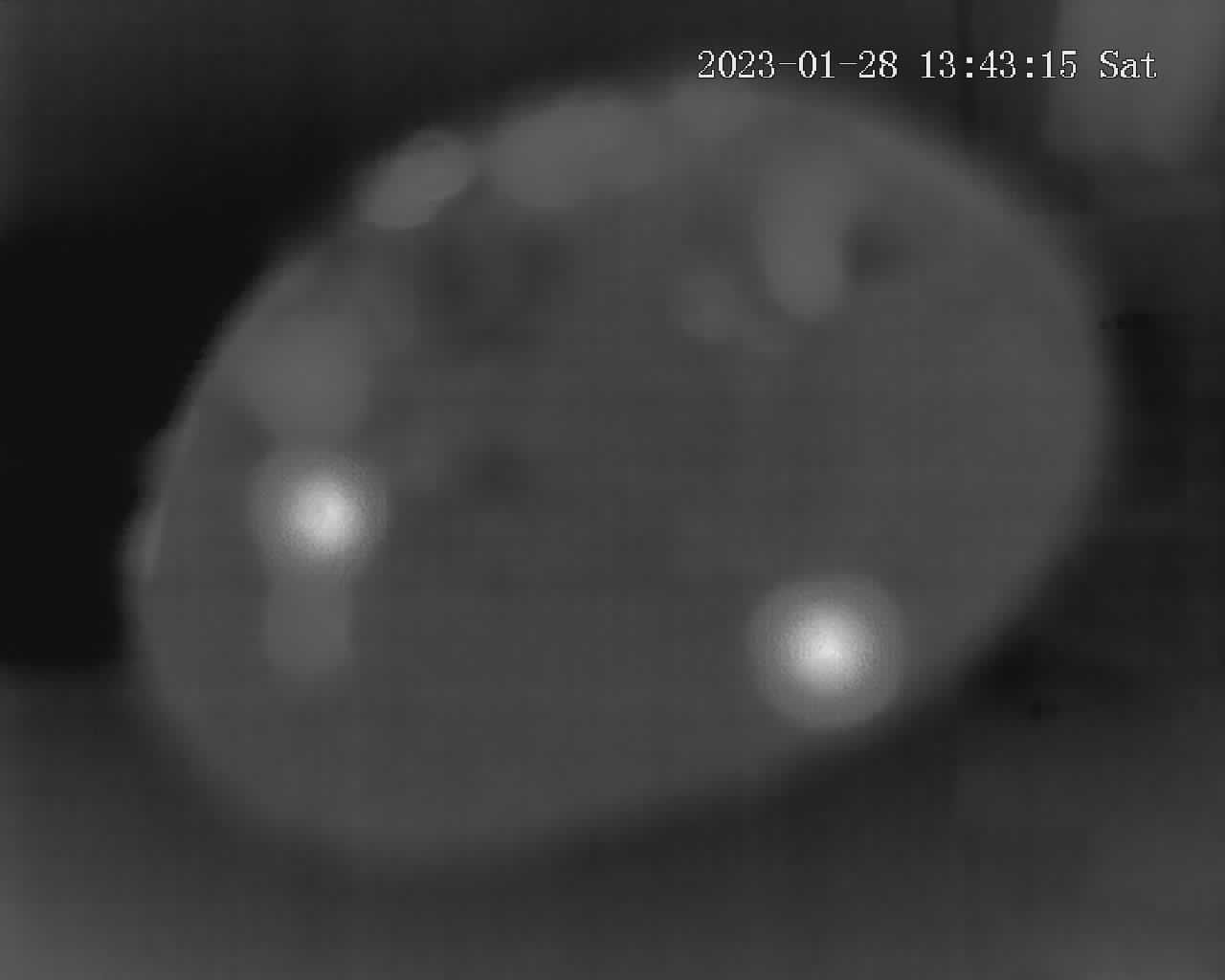}
        \includegraphics[scale=0.0733]{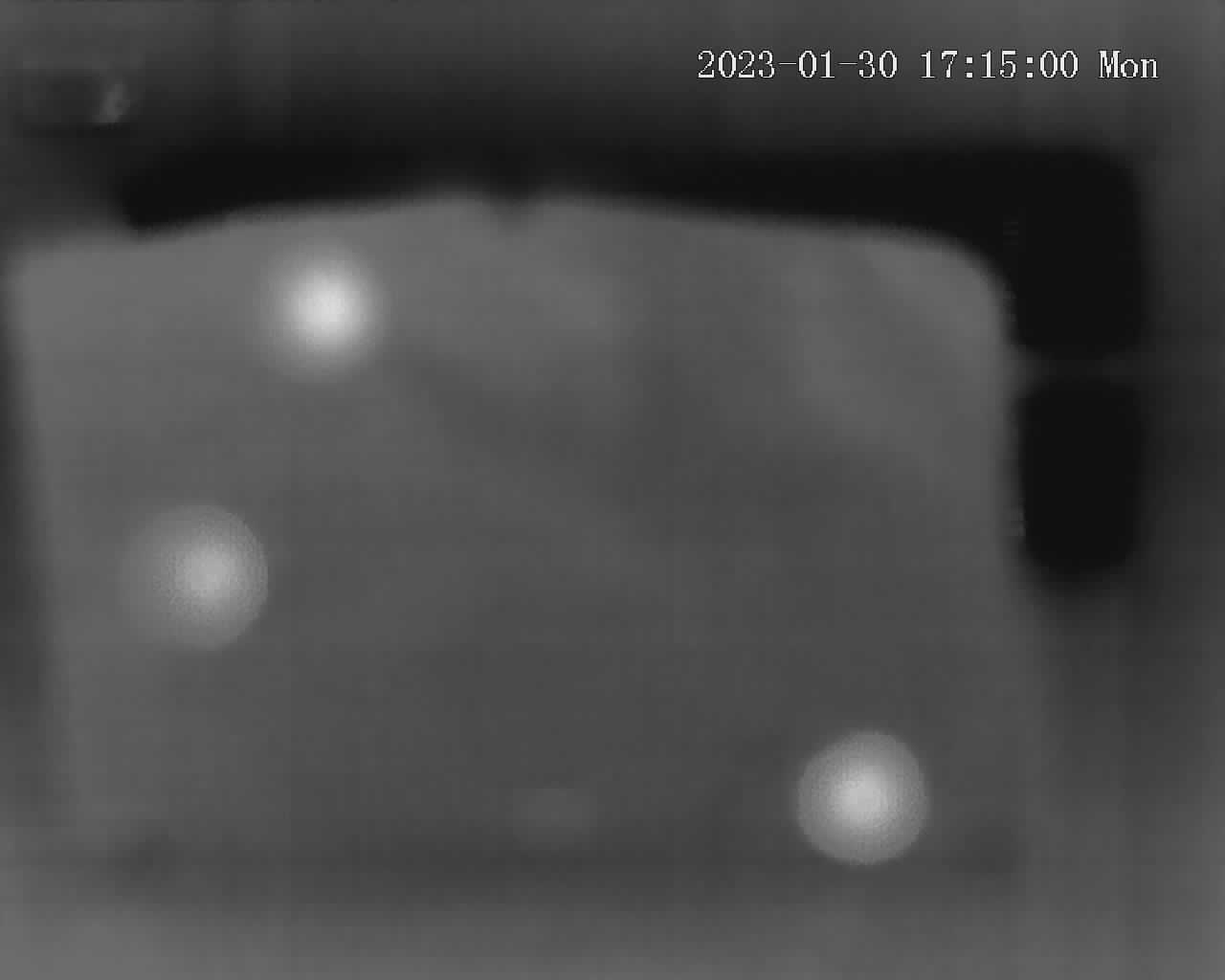}
        \includegraphics[scale=0.0733]{hm13.jpg}
        \includegraphics[scale=0.0733]{hm15.jpg}
        \caption{}
        \label{fig:ComparisonsInput}
    \end{subfigure}\par\medskip
    \begin{subfigure}{\textwidth}
        \centering
        \includegraphics[scale=0.146]{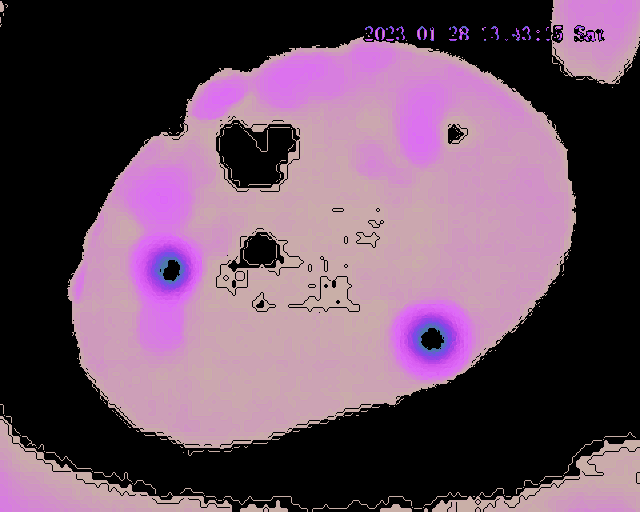}
        \includegraphics[scale=0.146]{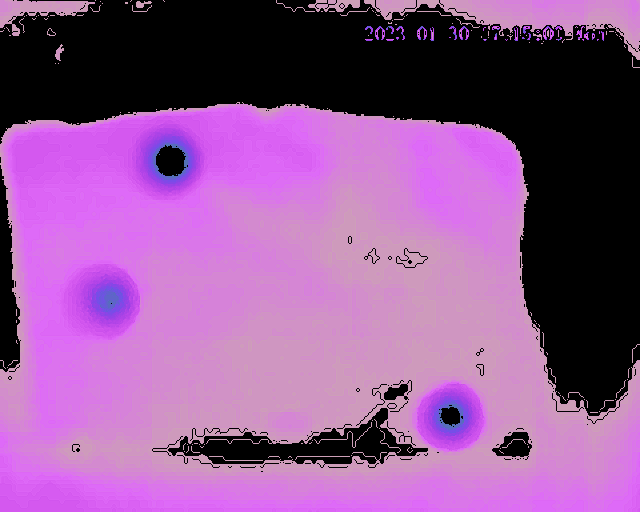}
        \includegraphics[scale=0.146]{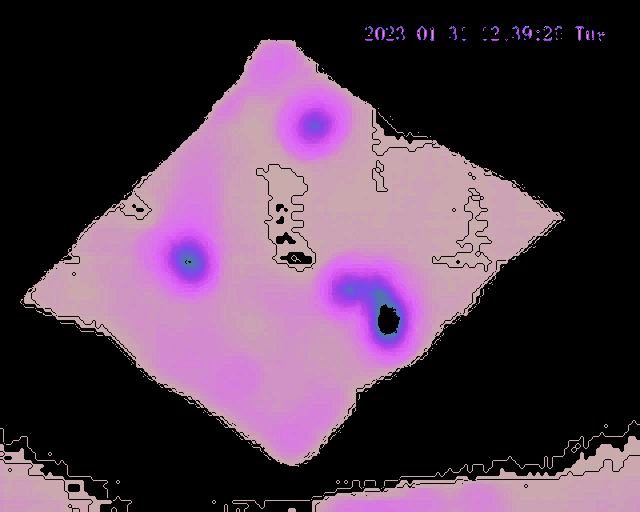}
        \includegraphics[scale=0.146]{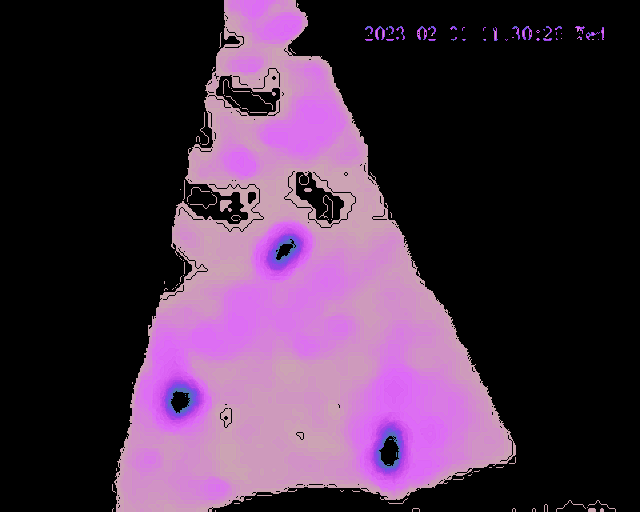}
        \caption{}
        \label{fig:ComparisonsKMeans}
    \end{subfigure}\par\medskip
    \begin{subfigure}{\textwidth}
        \centering
        \includegraphics[scale=0.146]{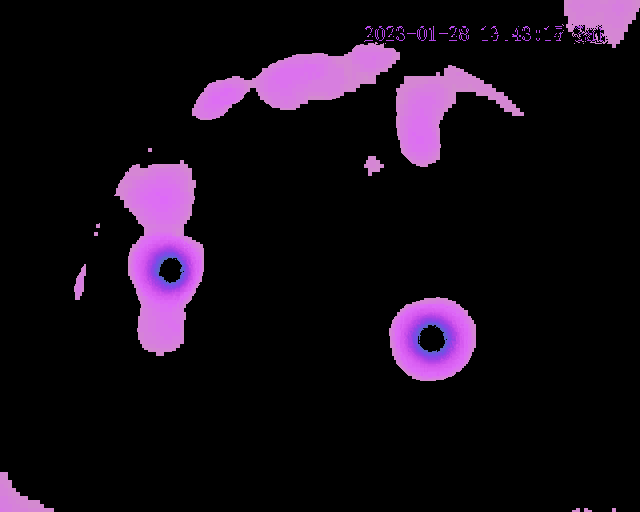}
        \includegraphics[scale=0.146]{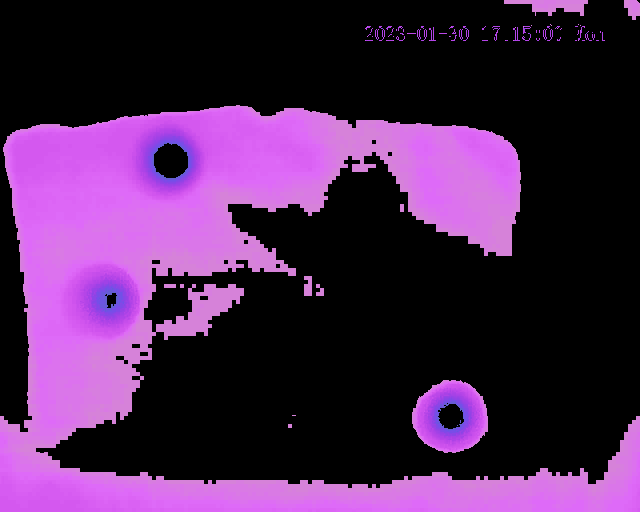}
        \includegraphics[scale=0.146]{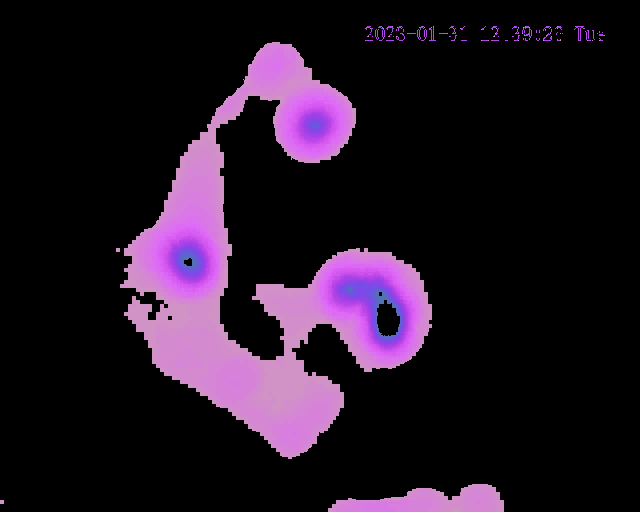}
        \includegraphics[scale=0.146]{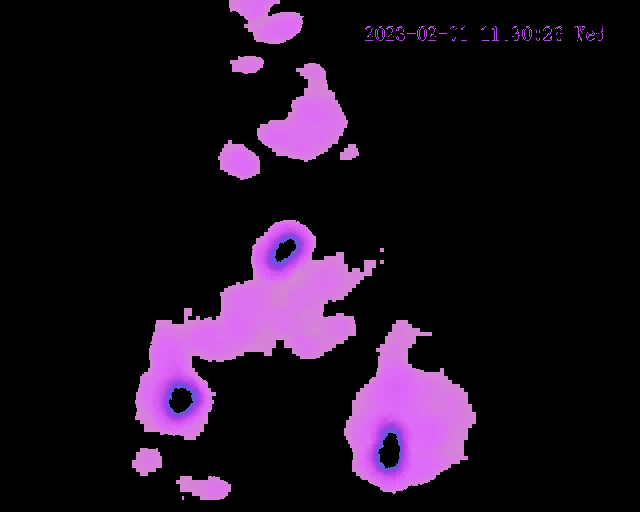}
        \caption{}
        \label{fig:ComparisonsKMeansPV}
    \end{subfigure}\par\medskip
    \begin{subfigure}{\textwidth}
        \centering
        \includegraphics[scale=0.146]{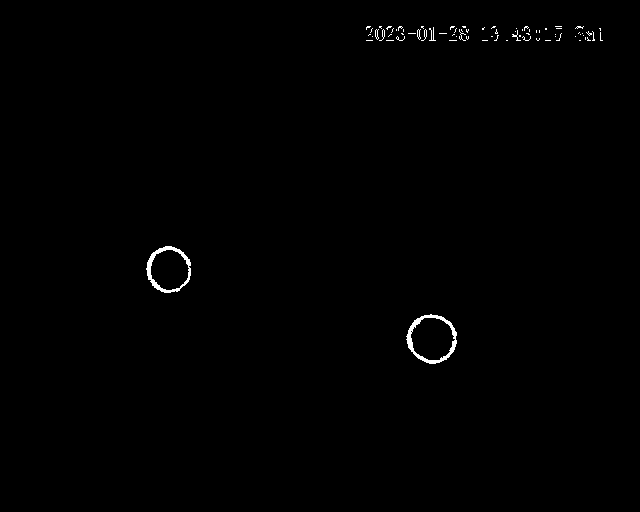}
        \includegraphics[scale=0.146]{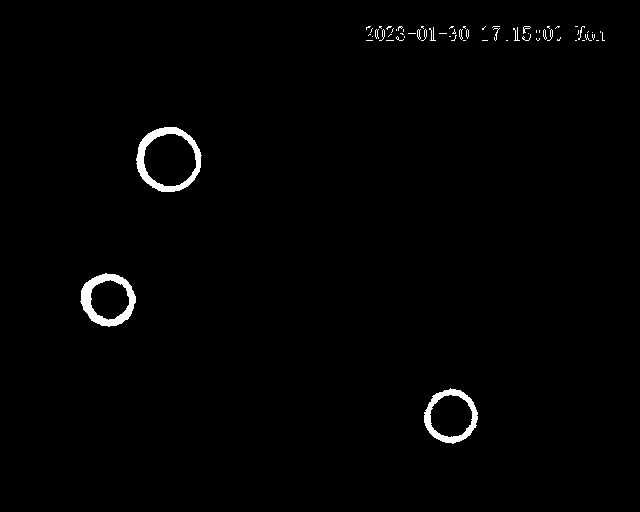}
        \includegraphics[scale=0.146]{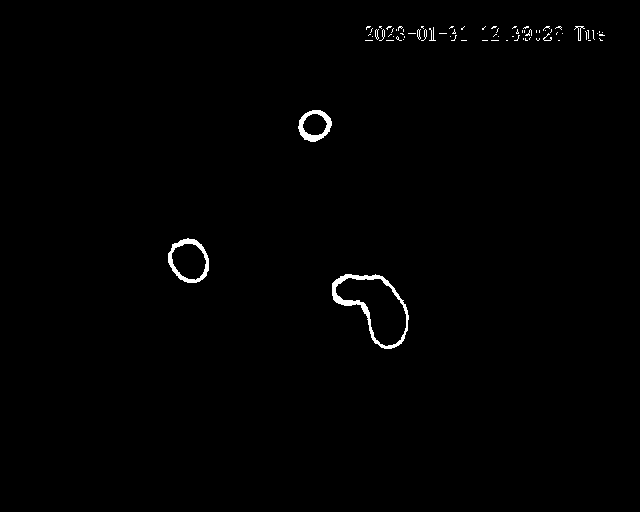}
        \includegraphics[scale=0.146]{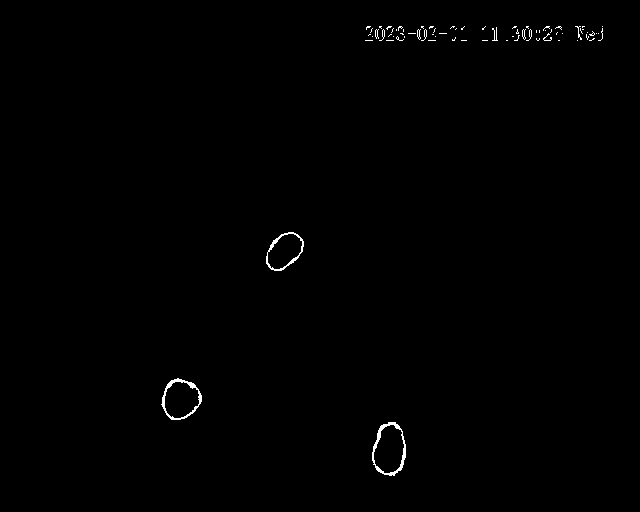}
        \caption{}
        \label{fig:ComparisonsHSV}
    \end{subfigure}
    \label{fig:Comparisons}
\end{figure}

\begin{figure}[htb]\ContinuedFloat
    \begin{subfigure}{\textwidth}
        \centering
        \includegraphics[scale=0.0726]{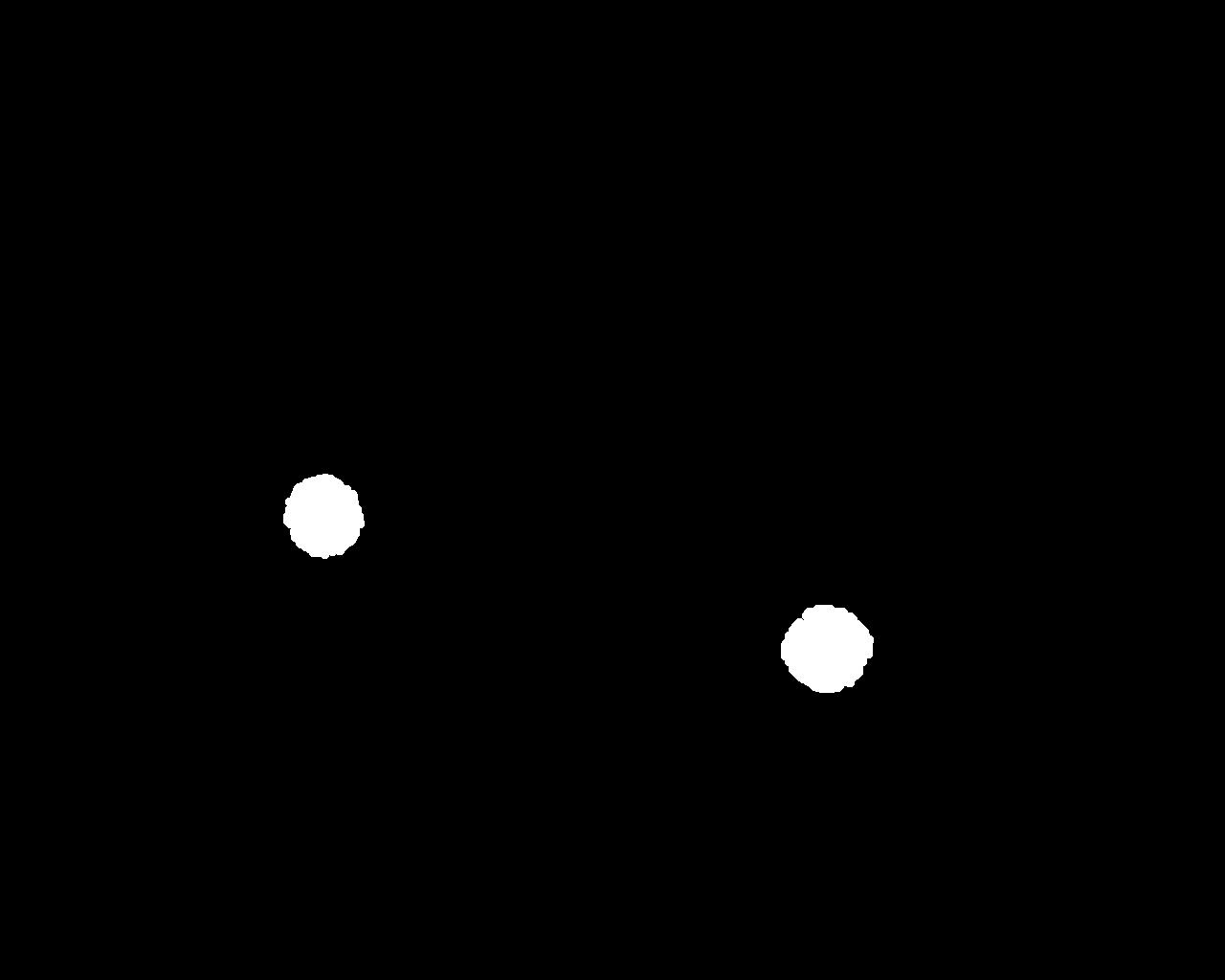}
        \includegraphics[scale=0.0726]{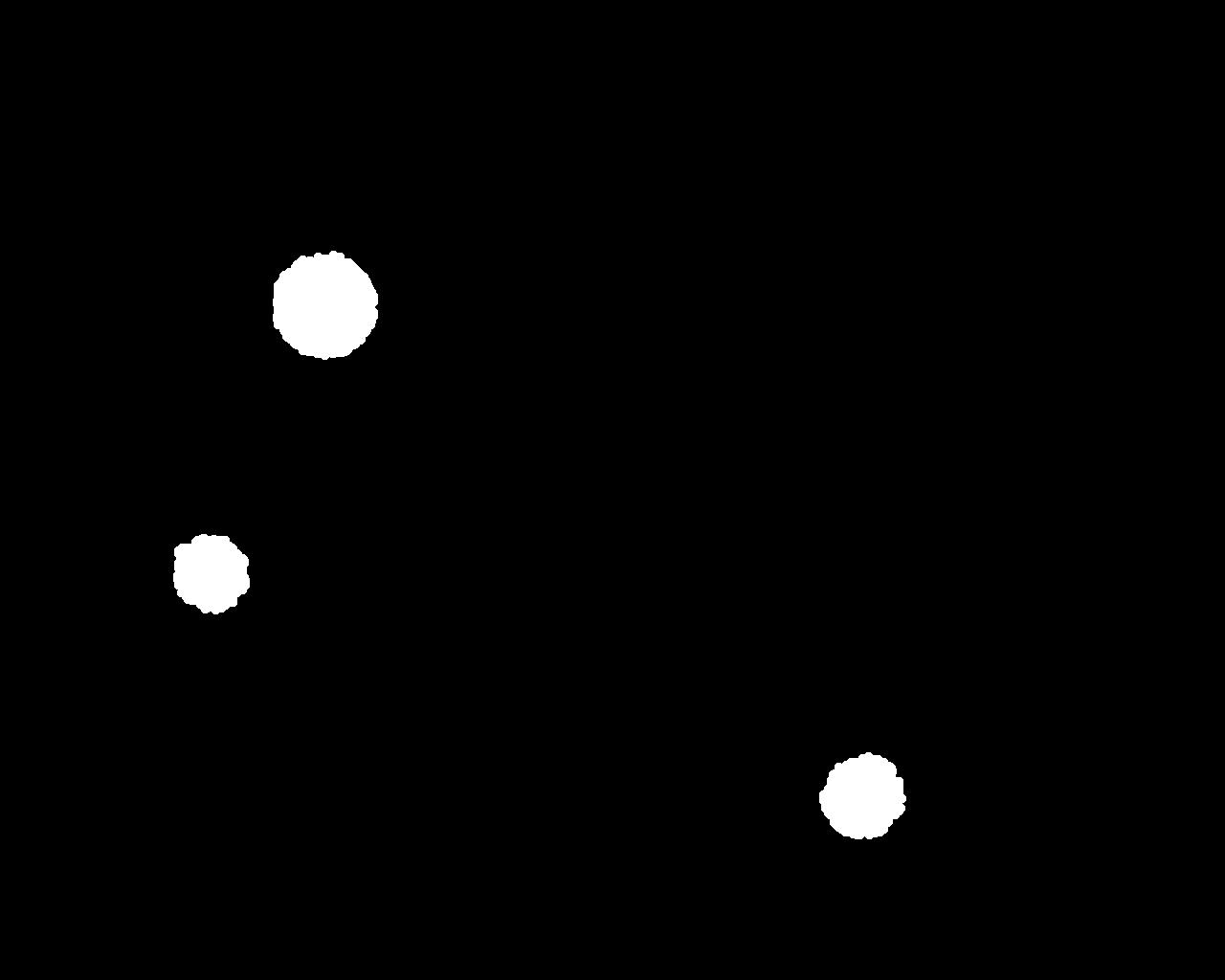}
        \includegraphics[scale=0.0726]{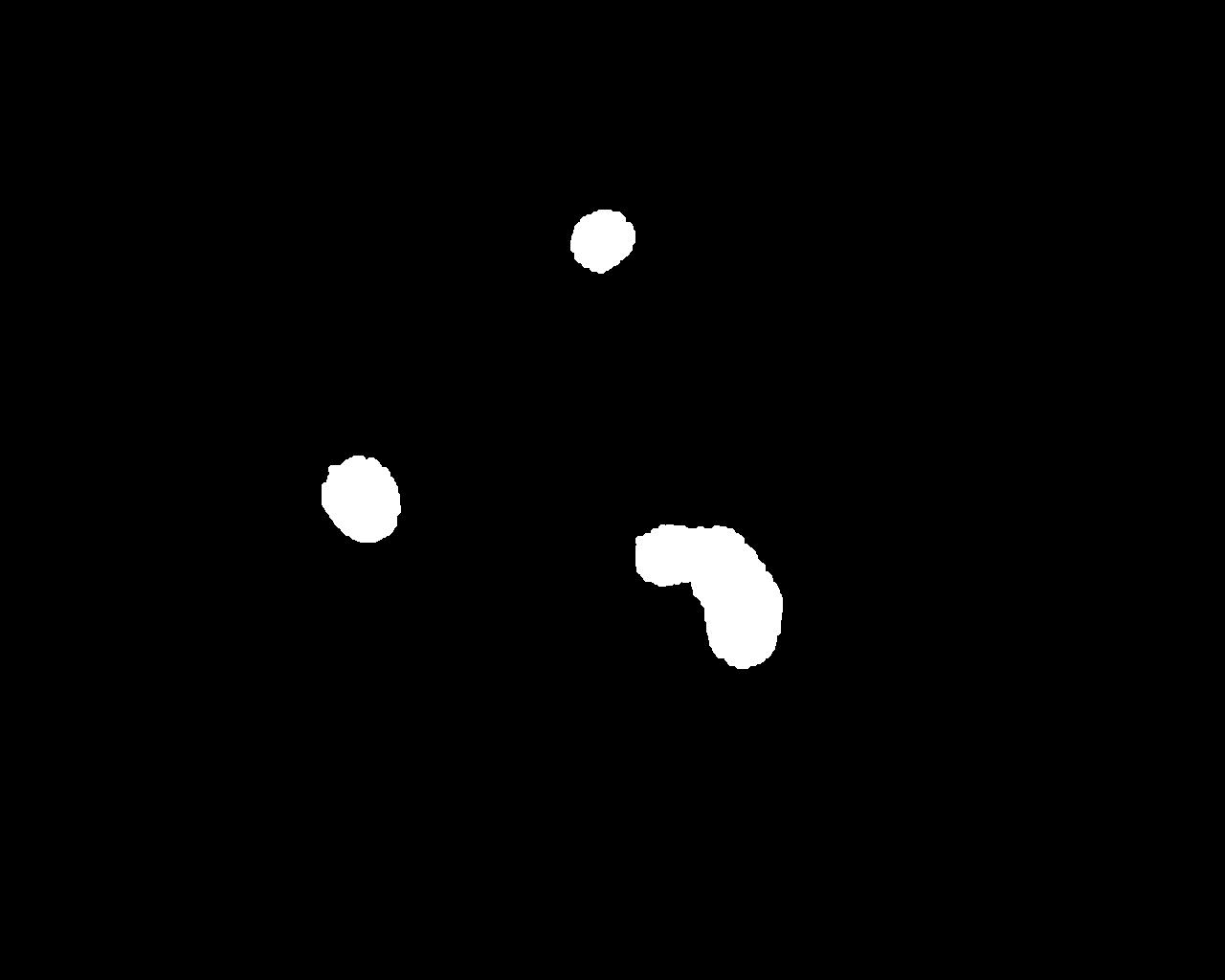}
        \includegraphics[scale=0.0726]{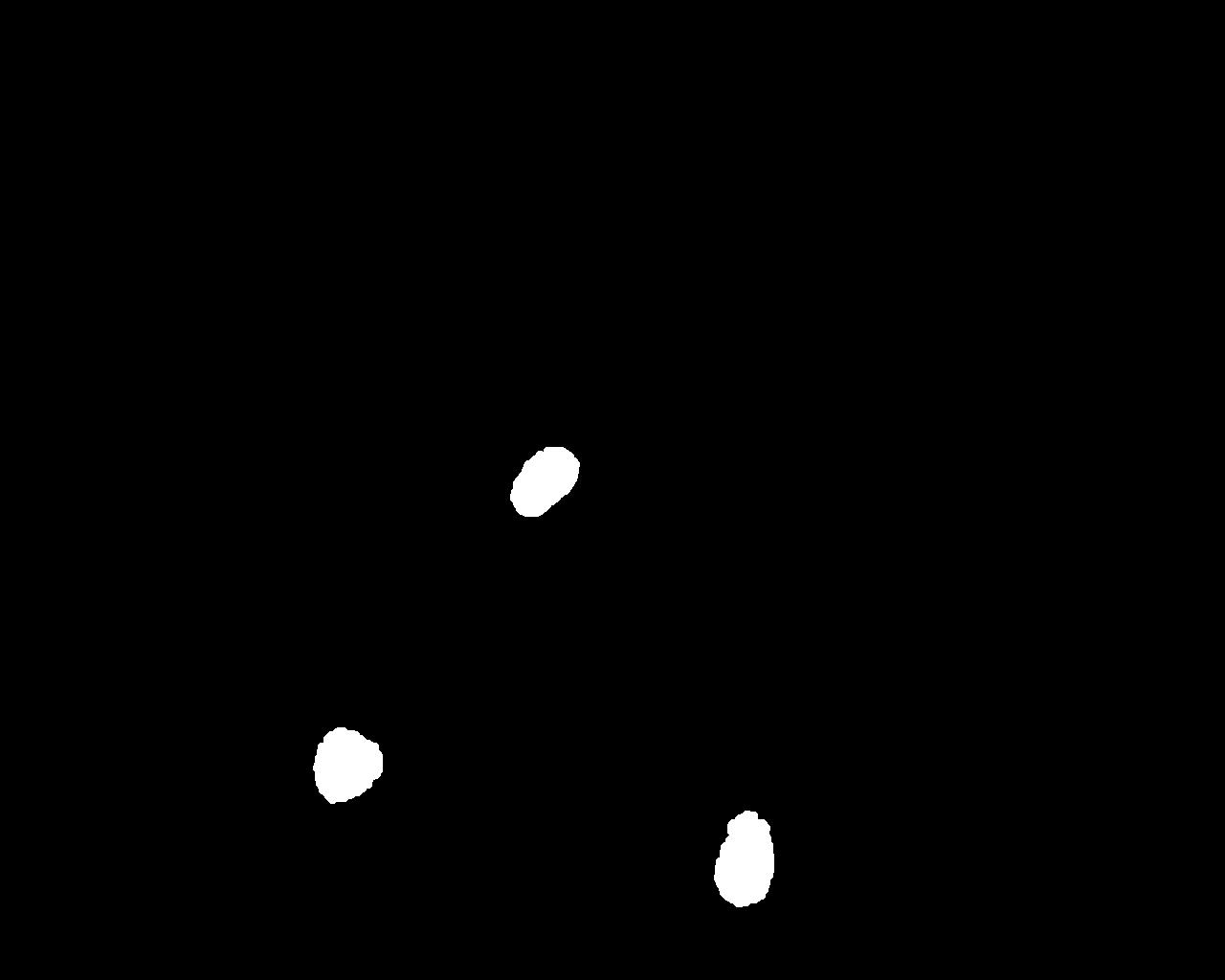}
        \caption{}
        \label{fig:ComparisonsOtsu}
    \end{subfigure}\par\medskip
    \begin{subfigure}{\textwidth}
        \centering
        \includegraphics[scale=0.28]{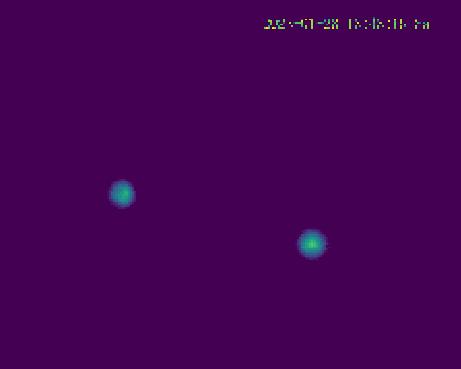}
        \includegraphics[scale=0.28]{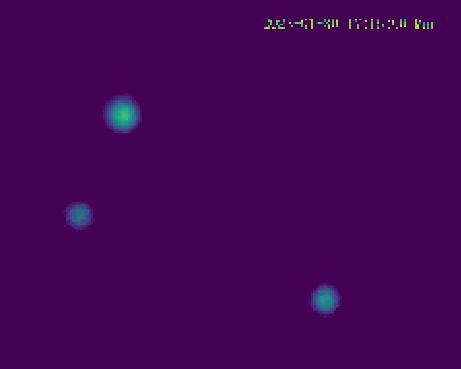}
        \includegraphics[scale=0.28]{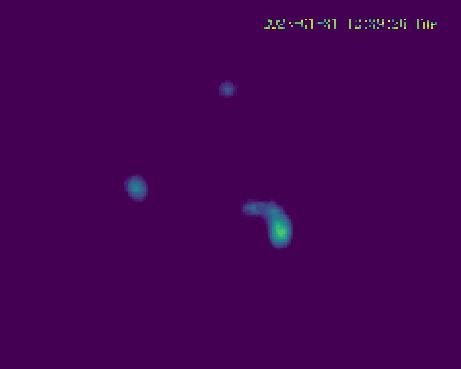}
        \includegraphics[scale=0.28]{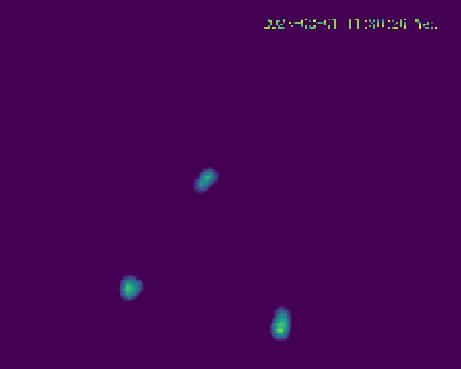}
        \caption{}
        \label{fig:ComparisonsOurs}
    \end{subfigure}
    \caption{A comparison between the result outputted by the existing hotspot identification methods. (a) shows the different input images on which the comparisons were conducted. The comparison has been conducted between the results of (b) K-Means segmentation \citep{mohd2017application}, (c) K-Means segmentation for photovoltaic hotspot detection \citep{salazar2016hotspots}, (d) HSV thresholding \citep{alajmi2019ir}, (e) Multi-level Otsu Segmentation \citep{afifah2021new}, and (f) our method for hotspot identification. Each column shows the results of the aforementioned techniques on a particular input image.}
\end{figure}

Comparison was also made with the approach proposed in \cite{salazar2016hotspots} for hotspot detection in photovoltaic cells. This work also utilizes the K-Means segmentation algorithm. Even though this K-Means method is able to isolate the multiple small hotspots, it is not a precise isolation, given that a lot of background is also present. Moreover, it relies on a lot of manual user input to narrow down on to the region containing the hotspots.  The outputs of this method can be observed in Figure \ref{fig:ComparisonsKMeansPV}. Furthermore, it was observed that this approach seems more suitable for thermal images containing large single hotspots. To test this, the comparison was also made with an input that has one large hotspot. Both, our method and the method proposed in \cite{salazar2016hotspots} outputted a precise identification of the single hotspot.
This comparison shows that our method is a more general technique, applicable to a wide variety of applications.

The technique in \cite{alajmi2019ir} was also reproduced for the purpose of comparison with our model. This method utilizes the HSV colour space and binary masks for the purpose of localization of the hotspots. However, the work depends on fixed specific threshold values that might not work for all cases, and therefore may not be generalizable. The outputs of this method can be observed in Figure \ref{fig:ComparisonsHSV}. It can be observed that even though the method is able to precisely identify the locations of the hotspots, it is only able to identify their boundaries. On the other hand, our method is able to identify the complete hotspots precisely (Figure \ref{fig:ComparisonsOurs}). It was also found that when we tune the suggested thresholds, we are able to identify the complete hotspots rather than just the boundaries. This upholds our thoughts about this method that using a fixed set of thresholds is not tractable for different image sets, and that threshold tuning will sometimes be required to adjust for particular thermal images for the identification of hotspots.

On comparison with the technique proposed in \cite{afifah2021new}, we found that even though this is a reasonable approach, it is not generalizable to all sets of thermal images of photovoltaic cells. This is because the heat signatures of different thermal images vary depending on the method used to capture them. Therefore, limiting the threshold number to given values is not the best approach, and the conclusion that higher the number of thresholds the more accurate the result does not apply every time. When testing the approach on our dataset, we empharically found the best thresholds; the results can be seen in Figure \ref{fig:ComparisonsOtsu}.

A quantitative comparison was also done of the methods mentioned above with our method by measuring the Dice coefficients from the results of each of the method. We randomly selected ten representative images from the TIHD dataset and the AAU VAP dataset for this purpose. The dice coefficient results have been summarized in Table \ref{tab:dicecoeff}. It can be seen from the results that our method was able to outperform the existing techniques for hotspot identification. \cite{afifah2021new} had suggested the use of 7,8 or 9 thresholds and concluded that higher number of thresholds leads to better results. However, this strategy did not work on the TIHD dataset, as the maximum number of thresholds that this method was able to use to segment the images from this dataset was 4. Any number higher than that failed to segment the image. Therefore, even though this was not suggested in \cite{afifah2021new}, the number of thresholds being used for this method in Table \ref{tab:dicecoeff} is 4, for the sake of comparisons. These results show that our method is highly precise when it comes to hotspot identification.
\begin{table}
    \centering
    \caption{Dice coefficient averages of the results outputted by the existing hotspot identification techniques - K-Means \citep{mohd2017application}, K-Means for photovoltaic panels \citep{salazar2016hotspots}, HSV Thresholding \citep{alajmi2019ir} and Multilevel Otsu \citep{afifah2021new}.}
    \begin{tabular}{cccccc}
        \hline
         & K-Means & K-Means 2 & HSV Thresh. & Otsu & Ours \\[0.5ex]
         \hline\hline
         AAU-VAP & $0.0768\pm0.06$ & $0.1390\pm0.05$ & $0.0159\pm0.01$ & $0.7423\pm0.08$ & $\textbf{0.7606}\pm\textbf{0.07}$ \\
         \hline
        TIHD & $0.1289\pm0.11$ & $0.2167\pm0.20$ & $0.2515\pm0.14$ & $0.5805\pm0.11$ & $\textbf{0.7359}\pm\textbf{0.10}$ \\
        \hline
    \end{tabular}
    
    \label{tab:dicecoeff}
\end{table} 

\section{Discussion and Conclusion}\label{conclusions}
In this project, a SimSiam inspired self-supervised learning model was created for the purpose of detecting hotspots for application in industrial settings where annotated datasets are scarce. Our method introduced a modified cost function for the network, introduced Xception network, and ensemble framework for adoptation to hotspots detection problem. 

A dataset THID consisting of 7361 thermal images was also created given the paucity of thermal image datasets. The model created was trained, and fine-tuned on a dataset for the purpose of classifying images as containing hotspots or 'normal'. The model so created was tested against existing supervised learning CNNs, as well as machine learning models, by training it on two sets of data (AAU-VAP trimodal dataset and TIHD dataset). It was shown experimentally, that the self-supervised model created performs in a competitive manner to the existing supervised CNNs, and can be considered a replacement to these models given its lower dependence on labelled datasets. 

An attempt was also made at model explainability to make the behaviour of the created model more transparent. This was done using GradCAM, which also helped us get the precisely isolate the detected hotspots. It was therefore shown that such a model is capable of being used in an industrial setting for the purpose of automation of hotspot detection to prevent equipment from reaching failure stages. Thereby preventing huge monetary as well as human life losses. With the use of this method, we waive the usual need for a large-scale, annotated dataset that is quite challenging to come across, as well as to create. This technique is a highly generalizable and accurate approach as well, with room for tweaks to be made for application of hotspot detection in different areas.

Potential future directions of research of this research could go into predictive maintenance of equipment along with condition monitoring. This could involve monitoring of the overall thermal patterns of an equipment over time, and predicting potential maintenance requirements to minimize downtime. Furthermore, with the use of some more data, the proposed approach can be further generalized in to detection of different types of possible defects other than hotspots. These could be defects like cracks, scratches, dents, etc. Creating such a system that is accurate when working with different scales of objects would further make the method more general, and increase its usability and potential applications in different fields. Further work can also be done in the direction of coming up with unique data augmentation techniques with specific suitability to thermal images that might improve the performance of supervised and self-supervised frameworks when working with thermal images.

\section{Acknowledgements}
This work was partly supported by a grant from Wonder Engineering Technologies Pte. Ltd., Singapore.

\bibliography{References}
\end{document}